\pdfoutput=1

\documentclass[10pt,twocolumn,letterpaper]{article}

\usepackage{wacv}

\usepackage{colortbl}  %
\usepackage{longtable} %
\graphicspath{{figures/}}
\definecolor{insample}{HTML}{D9D9D9}  %
\definecolor{samelab}{HTML}{EFEFEF}   %
\newcommand{\cmark}{\ensuremath{\checkmark}}
\newcommand{\xmark}{\ensuremath{\times}}

\newcommand{\dsname}{TrustUIQA\xspace}
\newcommand{\codeurl}{\url{https://github.com/TrustUIQA/TrustUIQA}\xspace}
\newcommand{\dataurl}{\url{https://huggingface.co/datasets/TrustUIQA/TrustUIQA}\xspace}
\newcommand{\naccTieBest}{0.72\xspace} %
\newcommand{\naccTieBestName}{LIQE\xspace} %
\newcommand{\nallFourNames}{ARNIQA-TID, MANIQA-PIPAL, NIMA-VGG16-AVA, QualiCLIP+-SPAQ, RALI\xspace} %
\newcommand{\nattWin}{100\%\xspace} %
\newcommand{\nblockSizes}{2, 3, 9 and 6\xspace} %
\newcommand{\ncleanMusiq}{33\%\xspace} %
\newcommand{\ndEMax}{51.2\xspace} %
\newcommand{\ndEMin}{24.7\xspace} %
\newcommand{\ndriftCropId}{044\xspace} %
\newcommand{\ndriftFrame}{1376\times768\xspace} %
\newcommand{\ndriftInW}{512\xspace} %
\newcommand{\neditSafeNames}{NMFC, VisualQuality-R1, IL-NIQE, UIQM, Q-Align, LIQE, ARNIQA, NR-UIQA\xspace} %
\newcommand{\neditorCost}{285\xspace} %
\newcommand{\nflipMaxHalf}{47\%\xspace} %
\newcommand{\nflipMaxHflip}{31\%\xspace} %
\newcommand{\nflipMedHalf}{27\%\xspace} %
\newcommand{\nflipMedHflip}{7\%\xspace} %
\newcommand{\nfusBeats}{10\xspace} %
\newcommand{\nfusNFeats}{68 to 72\xspace} %
\newcommand{\nfusOtherMax}{0.58\xspace} %
\newcommand{\nfusOtherMaxPair}{SOTA to SAUD2.0\xspace} %
\newcommand{\nfusPairs}{12\xspace} %
\newcommand{\nfusSaudBtoSaud}{0.63\xspace} %
\newcommand{\ngmadNSota}{72\xspace} %
\newcommand{\ngmadUciqeRankSota}{71\xspace} %
\newcommand{\ngridBeatAll}{30\%\xspace} %
\newcommand{\ngridBeatEdit}{39\%\xspace} %
\newcommand{\ngridCmp}{56\xspace} %
\newcommand{\ngridN}{20\xspace} %
\newcommand{\ngridWins}{30, 28, 24\xspace} %
\newcommand{\nhandDetectAny}{UIQM and UIConM\xspace} %
\newcommand{\nhuePhoto}{51\%\xspace} %
\newcommand{\nhueVote}{45\%\xspace} %
\newcommand{\nhumanChainTwenty}{1\%\xspace} %
\newcommand{\nhumanNoiseStrong}{9\%\xspace} %
\newcommand{\nidentLeft}{54\%\xspace} %
\newcommand{\nimplRhoMax}{0.79\xspace} %
\newcommand{\nimplRhoMin}{0.48\xspace} %
\newcommand{\ninterRaterMax}{0.54\xspace} %
\newcommand{\ninterRaterMed}{0.41\xspace} %
\newcommand{\ninterRaterMin}{0.11\xspace} %
\newcommand{\nkBestNameSaud}{UIQAVSI\xspace} %
\newcommand{\nkBestNameStudy}{VisualQuality-R1\xspace} %
\newcommand{\nkBestNameUidA}{QUIQ\xspace} %
\newcommand{\nkBestSaud}{0.92\xspace} %
\newcommand{\nkBestStudy}{0.44\xspace} %
\newcommand{\nkBestUidA}{0.70\xspace} %
\newcommand{\nkDisjointDropMax}{0.02\xspace} %
\newcommand{\nkInMax}{0.92\xspace} %
\newcommand{\nkInMin}{0.60\xspace} %
\newcommand{\nkLabMax}{0.65\xspace} %
\newcommand{\nkLabMin}{0.58\xspace} %
\newcommand{\nkMedSaud}{0.25\xspace} %
\newcommand{\nkMedSaudB}{0.21\xspace} %
\newcommand{\nkMedSota}{0.10\xspace} %
\newcommand{\nkMedStudy}{0.21\xspace} %
\newcommand{\nkMedUidA}{0.16\xspace} %
\newcommand{\nkOrderMax}{0.11\xspace} %
\newcommand{\nkOrderMedMax}{0.04\xspace} %
\newcommand{\nkOrderRankMin}{0.97\xspace} %
\newcommand{\nkOtherMax}{0.31\xspace} %
\newcommand{\nkSymWorstBest}{0.26\xspace} %
\newcommand{\nkSymWorstBestName}{Q-Align\xspace} %
\newcommand{\nkUciqeImplDiff}{0.16\xspace} %
\newcommand{\nkUciqeMatStudy}{0.02\xspace} %
\newcommand{\nkUciqeMatUidA}{0.34\xspace} %
\newcommand{\nkUciqeSaud}{$-$0.08\xspace} %
\newcommand{\nkUciqeSaudB}{$-$0.21\xspace} %
\newcommand{\nkUciqeSota}{$-$0.17\xspace} %
\newcommand{\nkUciqeUidA}{0.18\xspace} %
\newcommand{\nkUiqmSaudB}{0.34\xspace} %
\newcommand{\nkUiqmUidA}{0.18\xspace} %
\newcommand{\nkWorstBest}{0.23\xspace} %
\newcommand{\nkWorstBestName}{Q-Align\xspace} %
\newcommand{\nkWorstSecond}{0.22\xspace} %
\newcommand{\nkWorstSecondName}{NR-UIQA\xspace} %
\newcommand{\nnAllFive}{72\xspace} %
\newcommand{\nnAllFour}{5\xspace} %
\newcommand{\nnAttMetrics}{5\xspace} %
\newcommand{\nnAttScenes}{50\xspace} %
\newcommand{\nnAttScenesAll}{150\xspace} %
\newcommand{\nnBlockBig}{9\xspace} %
\newcommand{\nnBlockSmall}{2\xspace} %
\newcommand{\nnChoices}{1,920\xspace} %
\newcommand{\nnConditions}{12\xspace} %
\newcommand{\nnDetectNbEnh}{26\xspace} %
\newcommand{\nnDetectNbRep}{11\xspace} %
\newcommand{\nnDetectProEnh}{19\xspace} %
\newcommand{\nnDetectProRep}{22\xspace} %
\newcommand{\nnDriftImages}{2,800\xspace} %
\newcommand{\nnDriftImg}{20\xspace} %
\newcommand{\nnDriftMetrics}{72\xspace} %
\newcommand{\nnDriftScores}{73\xspace} %
\newcommand{\nnEditHalfSaud}{42\xspace} %
\newcommand{\nnEditHalfSota}{53\xspace} %
\newcommand{\nnEditHalfUidA}{54\xspace} %
\newcommand{\nnEditMetrics}{71\xspace} %
\newcommand{\nnEditSafe}{8\xspace} %
\newcommand{\nnEditScenesSaud}{100\xspace} %
\newcommand{\nnEditScenesSota}{200\xspace} %
\newcommand{\nnEditScenesUidA}{60\xspace} %
\newcommand{\nnEditorImages}{2,800\xspace} %
\newcommand{\nnGeneral}{57\xspace} %
\newcommand{\nnHandUW}{7\xspace} %
\newcommand{\nnImgMOS}{12,360\xspace} %
\newcommand{\nnLastDownAll}{24\xspace} %
\newcommand{\nnLastDownProRep}{38\xspace} %
\newcommand{\nnLearnedUW}{7\xspace} %
\newcommand{\nnMetrics}{73\xspace} %
\newcommand{\nnRankUidA}{73\xspace} %
\newcommand{\nnRaterPairs}{55\xspace} %
\newcommand{\nnRaterPairsBig}{36\xspace} %
\newcommand{\nnRaters}{20\xspace} %
\newcommand{\nnRatersAll}{20\xspace} %
\newcommand{\nnRejected}{7\xspace} %
\newcommand{\nnRejectedPhoto}{6\xspace} %
\newcommand{\nnRejectedSure}{4\xspace} %
\newcommand{\nnRobust}{72\xspace} %
\newcommand{\nnScenesMOS}{1,160\xspace} %
\newcommand{\nnShareSaudSaudB}{33\xspace} %
\newcommand{\nnSigmaUp}{76 of 80\xspace} %
\newcommand{\nnSota}{72\xspace} %
\newcommand{\nnSotaLowHalf}{61\xspace} %
\newcommand{\nnStudyImg}{32\xspace} %
\newcommand{\nnTopHitUidA}{3\xspace} %
\newcommand{\nnUciqeBelowChance}{3\xspace} %
\newcommand{\nnUciqeTauNeg}{3\xspace} %
\newcommand{\nnVLM}{2\xspace} %
\newcommand{\nnVotesPerCond}{160\xspace} %
\newcommand{\nnanMax}{38\%\xspace} %
\newcommand{\nnanMin}{27\%\xspace} %
\newcommand{\nnoiseWinFdum}{86\%\xspace} %
\newcommand{\nnoiseWinQuiq}{74\%\xspace} %
\newcommand{\nnoiseWinUciqe}{72\%\xspace} %
\newcommand{\nperImgBestNameNbRep}{ARNIQA-TID\xspace} %
\newcommand{\nperImgBestNameProRep}{ARNIQA-TID and RALI\xspace} %
\newcommand{\nperImgBestNbRep}{65\%\xspace} %
\newcommand{\nperImgBestProRep}{95\%\xspace} %
\newcommand{\npsnrMeanEight}{34.2 to 36.4\xspace} %
\newcommand{\npsnrMeanTwo}{44.8 to 47.0\xspace} %
\newcommand{\npsnrMinTwo}{43.8\xspace} %
\newcommand{\nraliToMusiq}{61\%\xspace} %
\newcommand{\nrepeatSame}{75\%\xspace} %
\newcommand{\nsigmaRatioMax}{2.8\xspace} %
\newcommand{\nsigmaRatioMin}{1.4\xspace} %
\newcommand{\nsigmaRatioProRep}{2.5\xspace} %
\newcommand{\nsurePhotoMax}{18\%\xspace} %
\newcommand{\ntauUciqeSaud}{$-$0.02\xspace} %
\newcommand{\ntauUciqeSaudB}{$-$0.18\xspace} %
\newcommand{\ntauUciqeSota}{$-$0.24\xspace} %
\newcommand{\ntauUciqeUidA}{0.23\xspace} %
\newcommand{\ntauUiqmSota}{0.42\xspace} %
\newcommand{\ntauUiqmUidA}{0.27\xspace} %
\newcommand{\ntwoafcAtCeiling}{4\xspace} %
\newcommand{\ntwoafcBest}{0.70\xspace} %
\newcommand{\ntwoafcBestName}{Q-Align\xspace} %
\newcommand{\ntwoafcCeiling}{0.69\xspace} %
\newcommand{\ntwoafcMedian}{0.60\xspace} %
\newcommand{\ntwoafcUciqe}{0.44\xspace} %
\newcommand{\nuciqeChainTwenty}{100\%\xspace} %
\newcommand{\nuciqeDNbEnh}{+0.25\xspace} %
\newcommand{\nuciqeDNbRep}{+0.36\xspace} %
\newcommand{\nuciqeDProEnh}{+0.26\xspace} %
\newcommand{\nuciqeDProRep}{+0.24\xspace} %
\newcommand{\nuciqeDarkId}{039\xspace} %
\newcommand{\nuciqeDarkShare}{58\%\xspace} %
\newcommand{\nuciqeDarkVal}{2.39\xspace} %
\newcommand{\nuciqeMatDNbEnh}{+0.29\xspace} %
\newcommand{\nuciqeMatDProRep}{+0.57\xspace} %
\newcommand{\nuciqeMatSSCMax}{90\%\xspace} %
\newcommand{\nuciqeMatSSCMin}{35\%\xspace} %
\newcommand{\nuciqeNoiseStrong}{100\%\xspace} %
\newcommand{\nuciqeOnRejected}{91\%\xspace} %
\newcommand{\nuciqePerImgProEnh}{5\%\xspace} %
\newcommand{\nuciqeSSCMax}{95\%\xspace} %
\newcommand{\nuciqeSSCSaud}{86\%\xspace} %
\newcommand{\nuciqeSSCSota}{95\%\xspace} %
\newcommand{\nuciqeSSCUidA}{60\%\xspace} %
\newcommand{\nuiqmOnRejected}{41\%\xspace} %
\newcommand{\nvqAccTie}{0.69\xspace} %
\newcommand{\nvqAccTieRank}{5\xspace} %
\newcommand{\nvqBelowOne}{4\%\xspace} %
\newcommand{\nvqStudyTie}{22\%\xspace} %
\newcommand{\nvqTieMax}{22\%\xspace} %
\newcommand{\nvqTieMin}{20\%\xspace} %
\newcommand{\nweAboveZero}{11\xspace} %
\newcommand{\nweBest}{0.15\xspace} %
\newcommand{\nweBestName}{LIQE\xspace} %
\newcommand{\nweMedEdit}{0.22\xspace} %
\newcommand{\nweMedK}{0.21\xspace} %
\newcommand{\nweN}{72\xspace} %
\newcommand{\nweOracle}{0.49\xspace} %
\newcommand{\nweRaterMed}{0.13\xspace} %
\newcommand{\nweSecond}{0.12\xspace} %
\newcommand{\nweSecondName}{Q-Align\xspace} %
\newcommand{\nweUciqe}{$-$0.03\xspace} %
\newcommand{\supp}{the appendix\xspace} %

\definecolor{wacvblue}{rgb}{0.21,0.49,0.74}
\usepackage[breaklinks,colorlinks,allcolors=wacvblue]{hyperref}

\def\wacvPaperID{*****} %
\def\confName{WACV}
\def\confYear{2027}

\title{Do Underwater Image Quality Metrics Agree with Human Perception?\\A Benchmark with Human Choices, Simple Edits and Generative Drift}

\author{Rishii Kumar Singh\qquad Abhinav Rajput\qquad Dinesh Kumar Vishwakarma\\
Delhi Technological University, India\\
{\footnotesize\nolinkurl{rishiikumarsingh_ene_25b18073@dtu.ac.in}, \nolinkurl{abhinavrajput_mc24b04_007@dtu.ac.in}, \nolinkurl{dinesh@dtu.ac.in}}
}

\begin{document}
\raggedbottom
\maketitle
\AddToShipoutPicture*{\AtPageLowerLeft{\put(\LenToUnit{0.75in},\LenToUnit{0.35in}){\parbox{\dimexpr\paperwidth-1.5in\relax}{\scriptsize \copyright~2027 IEEE. Personal use of this material is permitted. Permission from IEEE must be obtained for all other uses, in any current or future media, including reprinting/republishing this material for advertising or promotional purposes, creating new collective works, for resale or redistribution to servers or lists, or reuse of any copyrighted component of this work in other works.}}}}
\begin{abstract}
Underwater image enhancement is challenging to evaluate due to the lack of clean reference images, so researchers rely on no-reference image quality assessment (NR-IQA) metrics.
However, it remains unclear whether these metrics agree with human judgments.
We evaluate \nnMetrics hand-crafted underwater, learned underwater, vision-language and general-purpose metrics against human preferences on five references: four published opinion-score sets and a new paired-comparison study.
For each metric, we measure how often it selects the same image as people when comparing versions of the same scene.
We further test whether simple image edits and small adversarial perturbations raise metric scores above the enhancement people preferred, and whether scores decrease as repeated generative edits add noise.
Our results show that no metric consistently matches human preferences.
The best worst-case Cohen's $\kappa$ over the five references is only \nkWorstBest (\nkWorstBestName), while two raters in our study agree with a median $\kappa$ of \ninterRaterMed.
The widely used UCIQE often favors edits that people reject. On a Nano Banana Pro replica chain that should leave the image unchanged, only \nnDetectProRep of \nnDriftScores scores decrease at every checkpoint, while noise grows by $\times$\nsigmaRatioProRep.
We also introduce \dsname, a benchmark that combines the human references, our study choices, the edited, attacked and generated images, and the scores of every metric on every image.
The code is available at \codeurl and \dsname at \dataurl.
\end{abstract}

\section{Introduction}
\label{sec:intro}

\begin{figure*}[t]
\centering
\includegraphics[width=\textwidth]{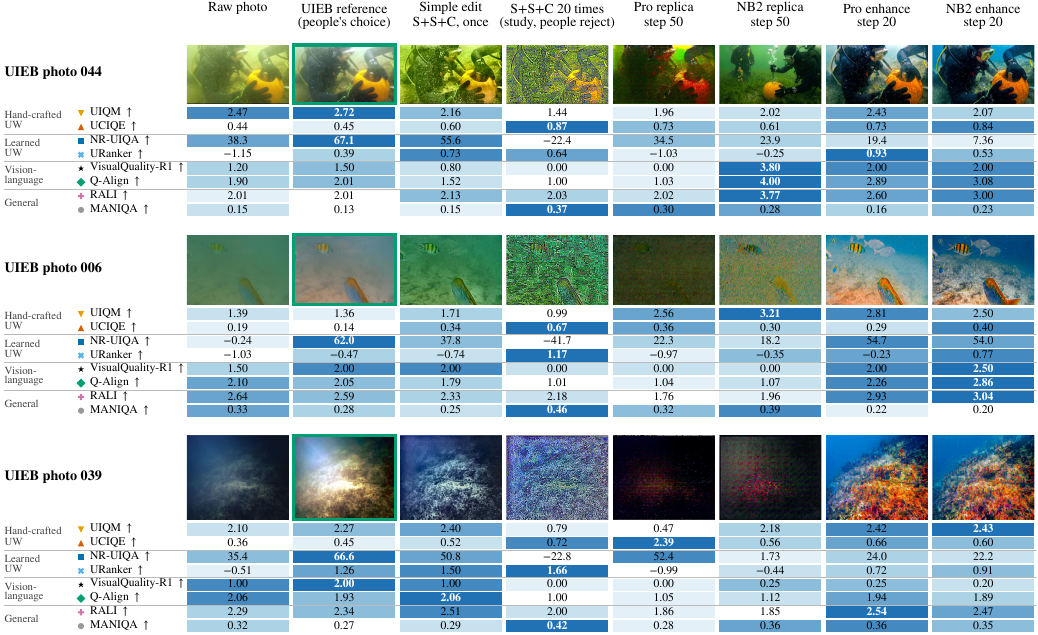}
\caption{\textbf{Metrics often score a simple edit or a repeatedly generated copy above the enhancement chosen by people.}
Each row block shows one UIEB~\cite{li2019underwater} photo: the raw image, the reference enhancement chosen by UIEB's volunteers (green frame), one saturate, sharpen and CLAHE~\cite{zuiderveld1994contrast} edit at moderate strength (S+S+C), a milder S+S+C step applied 20 times (a condition people reject in our study, \cref{sec:gaming}), and the last frame of the four Nano Banana chains of \cref{sec:drift}.
Below each photo are the scores of eight metrics, two per group: the default checkpoints with the highest and the lowest mean within-scene $\kappa$ with people in that group (\cref{sec:agree}).
$\uparrow$: a higher score means better quality, for all eight metrics.
Cell shade: rank of the panel among the eight for that metric, darkest and bold for the highest.
We show the three of the \ngridN drift photos in which a non-reference panel outscores the reference most often (\ngridWins of \ngridCmp comparisons).
Over all \ngridN photos, the rate is \ngridBeatAll, and \ngridBeatEdit for the single S+S+C edit.
\Cref{sec:agree,sec:gaming,sec:drift} measure these failures on full datasets.}
\label{fig:grid}
\end{figure*}

Underwater images lose red light, contrast and detail with depth, and many enhancement methods aim to restore them.
Since no clean reference exists for a real underwater image, these methods are compared using no-reference image quality assessment (NR-IQA) metrics.
UIQM~\cite{panetta2015human} and UCIQE~\cite{yang2015underwater} are the most common choices, learned metrics trained on underwater opinion scores are increasingly used~\cite{guo2023underwater,jiang2022underwater,yi2024no}, and general vision-language models such as Q-Align~\cite{wu2023q} can also score underwater images.
In each case, a higher score is interpreted as a better image.
The same assumption underlies other restoration tasks, where general NR-IQA metrics such as NIQE~\cite{mittal2012making} rank low-light enhancement~\cite{jiang2021enlightengan} and perceptual super-resolution~\cite{blau20182018} methods.

This interpretation holds only if the metric orders images as people do, and \cref{fig:grid} shows cases where it does not.
In all three photos, UCIQE assigns its highest score to the S+S+C edit applied 20 times or to the 50th Pro replica frame, and URanker~\cite{guo2023underwater} and MANIQA~\cite{yang2022maniqa} score the 20-times edit above the reference, whereas NR-UIQA~\cite{yi2024no} ranks the reference first.
Since a few examples are not conclusive, this paper measures how often such reversals occur.
We evaluate \nnMetrics metrics (\nnHandUW hand-crafted underwater, \nnLearnedUW learned underwater, \nnVLM vision-language and \nnGeneral general-purpose) against five human references and address four questions.
(1)~Do metrics choose between enhancements of the same photo as people do?
(2)~Do learned underwater metrics generalize to data from other labs?
(3)~Can simple edits or small gradient steps raise a score above the enhancement people preferred?
(4)~Do metrics detect the noise and color shifts introduced by repeated generative edits?
Our results show that the answer to (1) and (2) is mostly no and the answer to (3) is yes.
For (4), most scores miss these changes under the strict rule of Banana100~\cite{tang2026banana100}, and some, including UCIQE, even rise.

Our main contributions are as follows:
\begin{itemize}[leftmargin=*,itemsep=1pt,topsep=2pt]
\item \textbf{A within-scene agreement evaluation of \nnMetrics metrics on five human references} (\cref{sec:bench,sec:agree}): four published opinion-score sets (\nnImgMOS images, \nnScenesMOS scenes) and a new paired-comparison study with \nnRaters raters and \nnChoices choices. The best worst-case $\kappa$ is only \nkWorstBest (\nkWorstBestName). The five learned underwater metrics with a known training set reach \nkInMin to \nkInMax on it, but at most \nkOtherMax on data from other labs.
\item \textbf{Two score manipulation tests} (\cref{sec:gaming}). A saturate, sharpen and CLAHE filter outscores the best human-rated enhancement in \nuciqeSSCSota of SOTA~\cite{wang2022generation} scenes under UCIQE. In our study, UCIQE prefers the edit in \nuciqeOnRejected of pairs where most people rejected it. Ten PGD~\cite{madry2017towards} steps at $\epsilon=2/255$ win every attacked scene for all \nnAttMetrics differentiable metrics we attack.
\item \textbf{A generative drift test} (\cref{sec:drift}) with \nnDriftImages images from four chains of repeated Nano Banana Pro and Nano Banana~2~\cite{google2025nbpro,google2026nb2} edits. Noise increases by $\times$\nsigmaRatioMin to $\times$\nsigmaRatioMax. On the Pro replica chain, where the image should not change, only \nnDetectProRep of \nnDriftScores scores decrease at every checkpoint and \nnLastDownProRep by the last step, while UCIQE increases on every chain.
\item \textbf{\dsname}, a benchmark that combines the five human references with our study choices, the edited, attacked and generated images, and the scores and $\kappa$ tables of all \nnMetrics metrics, so that new metrics can be evaluated in the same way (\cref{sec:conclusion}).
\end{itemize}
In addition, \cref{sec:why} shows that the public UCIQE code returns no score for \nnanMin to \nnanMax of the opinion-score images and offers a hypothesis for why its formula rewards simple edits.
\Cref{sec:conclusion} summarizes the results as five recommendations for future evaluations.

\section{Related work}
\label{sec:related}

\paragraph{Underwater quality metrics and opinion-score sets.}
Hand-crafted underwater metrics combine color, sharpness and contrast measures: UIQM~\cite{panetta2015human}, UCIQE~\cite{yang2015underwater}, CCF~\cite{wang2018imaging} and FDUM~\cite{yang2021reference}.
Learned metrics are trained on opinion scores of enhanced underwater images: URanker~\cite{guo2023underwater}, NUIQ~\cite{jiang2022underwater}, Twice-Mixing~\cite{fu2022twice}, NMFC~\cite{chen2023underwater}, NR-UIQA~\cite{yi2024no}, UIQAVSI~\cite{chen2025underwater} and QUIQ~\cite{karimi2025quiq}.
The opinion-score sets they use, UID2021~\cite{hou2023uid2021}, SAUD~\cite{jiang2022underwater}, SAUD2.0~\cite{jiang2024toward} and SOTA~\cite{wang2022generation}, evaluate metrics by whole-set correlation with their scores. Other benchmarks avoid opinion scores by using color charts~\cite{berman2020underwater} or a detection task~\cite{liu2020real}. Li and Cavallaro~\cite{li2022limits} found that UIQM, UCIQE and CCF do not match perceived quality, Lin et al.~\cite{lin2022crowdsourced} reached a similar conclusion from a crowdsourced study of enhanced underwater images, and recent studies review and compare subjective and objective evaluation of underwater enhancement~\cite{shen2025underwater,palazari2026towards}.
We do not claim the first evidence of these limitations. Our contribution is the combination of five human references, including a new study of simple edits, with \nnMetrics metrics, simple-edit baselines, stress tests and implementation checks.

\paragraph{Testing quality metrics.}
gMAD~\cite{ma2016group} compares metrics on images where they disagree most, and later work identifies failures of blind models in the wild~\cite{wang2021troubleshooting}. BAPPS~\cite{zhang2018unreasonable} evaluates metrics by how often they match a two-alternative human choice, and Q-Bench~\cite{wu2024q} tests the low-level perception of vision-language models.
Optimizing images for a quality model can raise its score without improving the images~\cite{ding2021comparison}, small gradient attacks raise many NR-IQA scores~\cite{antsiferova2024comparing,zhang2022perceptual}, and defenses against such attacks have been benchmarked~\cite{gushchin2024guardians}.
Unlike these general studies, we test underwater images and metrics, add simple image edits as a gradient-free way to manipulate scores, and compare edits and attacks with the enhancement people preferred in underwater opinion-score sets.

\paragraph{Generative drift.}
Banana100~\cite{tang2026banana100} asks Nano Banana Pro to copy an image 100 times and finds that many NR-IQA metrics do not decrease as the copies degrade. An evaluation on 14 low-level tasks finds its outputs visually good but behind on reference-based metrics~\cite{zuo2025nano}.
Unlike Banana100, we run the test on underwater images with two editors and two prompts, an exact-copy prompt and an enhancement prompt, and include the hand-crafted and learned underwater metrics and RALI~\cite{zhao2026reasoning}.

\section{Benchmark}
\label{sec:bench}

\subsection{Human references}
\label{sec:refs}

\begin{table}[t]
\centering\small
\begin{tabular}{@{}lrrr@{}}
\toprule
Reference & Scenes & Methods & Images \\
\midrule
UID2021~\cite{hou2023uid2021} & 60 & 16 & 960 \\
SAUD~\cite{jiang2022underwater} & 100 & 10 & 1,000 \\
SAUD2.0~\cite{jiang2024toward} & 200 & 12 & 2,400 \\
SOTA~\cite{wang2022generation} & 800 & 10 & 8,000 \\
\midrule
Our study & 32 & 12 edits & 1,920 choices \\
\bottomrule
\end{tabular}

\caption{\textbf{Five human references.} The four opinion-score sets rate several enhancements of each underwater photo (a scene). Our study pairs a raw UIEB~\cite{li2019underwater} photo with one edit of it.}
\label{tab:refs}
\end{table}

\Cref{tab:refs} lists the references.
UID2021~\cite{hou2023uid2021}, SAUD~\cite{jiang2022underwater}, SAUD2.0~\cite{jiang2024toward} and SOTA~\cite{wang2022generation} assign a mean opinion score to every enhancement of every scene.
SAUD and SAUD2.0 come from the same lab and share \nnShareSaudSaudB photos, so we treat them as one group when evaluating generalization. Scene counts are per set, so these photos are counted twice in the \nnScenesMOS scenes.

\paragraph{Our paired study.}
The opinion-score sets contain only outputs of published enhancement methods.
To test edits that a metric may reward but people may not, we conducted a two-alternative forced-choice study on \nnStudyImg raw UIEB~\cite{li2019underwater} photos.
Each trial shows the raw photo and one edited version side by side in random order with the question ``Pick the one with better overall quality.''
The \nnConditions edits range from mild CLAHE~\cite{zuiderveld1994contrast}, contrast, sharpening and saturation to strong noise and a chain of the saturate, sharpen and CLAHE filters applied 5 or 20 times (\cref{sec:a_setup} of \supp lists all strengths).
Each rater saw one of four blocks of 8 photos, with 96 test trials plus 6 attention checks (a strongly blurred copy), 5 repeated pairs and 4 identical pairs.
All \nnRatersAll raters passed every attention check, which gives \nnVotesPerCond votes per edit and \nnChoices choices in total.
Raters who saw the same block agree with a median Cohen's $\kappa$ of \ninterRaterMed (range \ninterRaterMin to \ninterRaterMax over \nnRaterPairs pairs). We use this value as a reference level for metric agreement, not as an upper bound. Of the \nnRaterPairs rater pairs, \nnRaterPairsBig come from the block with \nnBlockBig raters.

\subsection{Metrics}
\label{sec:metrics}

We evaluate \nnMetrics metrics in four groups.
\emph{Hand-crafted underwater} (\nnHandUW): UIQM and its three terms UICM, UISM and UIConM~\cite{panetta2015human}, UCIQE~\cite{yang2015underwater}, CCF~\cite{wang2018imaging} and FDUM~\cite{yang2021reference}.
\emph{Learned underwater} (\nnLearnedUW): the seven metrics of \cref{sec:related}, using the authors' code and weights.
NUIQ~\cite{jiang2022underwater}, NR-UIQA~\cite{yi2024no}, UIQAVSI~\cite{chen2025underwater} and NMFC~\cite{chen2023underwater} were trained on SAUD, and QUIQ~\cite{karimi2025quiq} on UID2021.
URanker was trained on enhancements of all 890 UIEB raw photos~\cite{guo2023underwater}, which include the photos of our study and drift test.
\emph{Vision-language} (\nnVLM): Q-Align~\cite{wu2023q} and VisualQuality-R1~\cite{wu2026visualquality}.
VisualQuality-R1 uses the prompt from its model card with greedy decoding. Every answer contained a score, although \nvqBelowOne of the scores on the opinion-score sets fall below the nominal range of 1 to 5.
\emph{General-purpose} (\nnGeneral): the no-reference metrics in IQA-PyTorch~\cite{chen2022iqa}, several with more than one checkpoint, namely BRISQUE~\cite{mittal2012no}, NIQE~\cite{mittal2012making}, IL-NIQE~\cite{zhang2015feature}, PIQE~\cite{venkatanath2015blind}, NRQM~\cite{ma2017learning}, PI~\cite{blau20182018}, CNNIQA~\cite{kang2014convolutional}, WaDIQaM~\cite{bosse2017deep}, DBCNN~\cite{zhang2018blind}, HyperIQA~\cite{su2020blindly}, MetaIQA~\cite{zhu2020metaiqa}, NIMA~\cite{talebi2018nima}, PaQ-2-PiQ~\cite{ying2020patches}, MUSIQ~\cite{ke2021musiq}, UNIQUE~\cite{zhang2021uncertainty}, TReS~\cite{golestaneh2022no}, MANIQA~\cite{yang2022maniqa}, CLIP-IQA~\cite{wang2023exploring}, LIQE~\cite{zhang2023blind}, TOPIQ~\cite{chen2024topiq}, ARNIQA~\cite{agnolucci2024arniqa}, QualiCLIP~\cite{agnolucci2024quality}, the no-reference branch of A-FINE~\cite{chen2025toward}, MA-CLIP~\cite{liao2026beyond}, the LAION aesthetic predictor~\cite{schuhmann2024christophschuhmann,schuhmann2022laion} and image entropy, plus RALI~\cite{zhao2026reasoning}. LIQE-mix was scored only on UID2021, SAUD and SAUD2.0.
All scores are oriented so that higher is better, and lower-is-better scores such as NIQE are negated.
Some of the \nnMetrics metrics are additional checkpoints of the same model (eight for ARNIQA) or terms of UIQM, so the metric counts below give these models more weight.
UIQM follows the widely used FUnIE-GAN~\cite{islam2020fast} code with two fixes (the trimmed-mean index and the UISM blue weight), FDUM is a Python port of the authors' MATLAB code, and CCF is our implementation of the published formula.
UCIQE uses our NumPy port, which divides $L^*$ and chroma by 100 and takes the 1st to 99th percentile range of $L^*$ as contrast. \Cref{sec:why} compares it with the public MATLAB code.

\subsection{Agreement within a scene}
\label{sec:kappa}

An enhancement paper asks which of several outputs for the same photo is best.
We therefore compare each metric with people on pairs of enhancements of the same scene.
For each pair with different opinion scores, we check whether the metric prefers the same image as people, pool all pairs of a set into one $2\times2$ table and report Cohen's $\kappa$~\cite{cohen1960coefficient}.
$\kappa=0$ indicates the agreement expected by chance from the two sets of choices, and $\kappa=1$ full agreement.
In each pair, the image listed earlier in the set's file comes first. Counting both orders of every pair instead changes the median metric's $\kappa$ by at most \nkOrderMedMax on any set and no metric's $\kappa$ by more than \nkOrderMax, and it leaves the ranking of metrics almost unchanged (Spearman correlation of at least \nkOrderRankMin). The best worst-case $\kappa$ becomes \nkSymWorstBest (\nkSymWorstBestName) instead of \nkWorstBest, so the absolute values depend somewhat on this convention. When both orders are counted, pairwise accuracy equals $(\kappa+1)/2$, so accuracy and this order-free $\kappa$ rank metrics in the same way. Rankings between metrics are descriptive, and we do not test their differences for significance.
Pairs that the metric scores as equal are excluded. This removes \nvqTieMin to \nvqTieMax of the pairs for VisualQuality-R1, the only metric that ties more than 5\% of them, and \nvqStudyTie of the study pairs.
In our study, the first image is always the raw photo.
The 95\% intervals resample scenes 1,000 times~\cite{stine1989introduction}.
In our study, each vote is one pair, and the interval resamples raters and photos together.
We do not use whole-set rank correlation, because it also rewards a metric for distinguishing different photos, which no enhancement comparison requires.

\subsection{Stress tests}
\label{sec:stress}

\paragraph{Simple edits.} We apply six edits at moderate strength to the raw photo of each scene: CLAHE, a red gain, Gaussian noise, saturation plus unsharp masking (S+S), S+S followed by CLAHE (S+S+C) and S+S plus noise (S+S+N). We use all \nnEditScenesUidA UID2021 scenes, all \nnEditScenesSaud SAUD scenes and \nnEditScenesSota SOTA scenes, and count the scenes in which the edit scores above the enhancement with the highest opinion score.
\paragraph{Gradient attack.} For \nnAttMetrics differentiable metrics, we apply 10 PGD~\cite{madry2017towards} steps of size $\epsilon/4$ under an $L_\infty$ bound $\epsilon\in\{2/255, 8/255\}$ to the raw photo of the first \nnAttScenes scenes of the same three sets, save the result as an 8-bit PNG and rescore it. Since the photo is 8-bit and $255\epsilon$ is an integer, rounding keeps the change within the bound.
\paragraph{Generative drift.} Following Banana100~\cite{tang2026banana100}, we send \nnDriftImg UIEB photos (512\,px) to Nano Banana Pro (gemini-3-pro-image) and Nano Banana~2 (gemini-3.1-flash-image)~\cite{google2025nbpro,google2026nb2}, feed each output back as the next input with a fresh request, and keep the output size of the model. We score the chains with \nnDriftMetrics of the metrics (all but LIQE-mix) and the patched MATLAB UCIQE code, \nnDriftScores scores in total.
The replica prompt is Banana100's ``Produce an exact replica of the provided image, with no alterations.'' (50 steps).
The enhance prompt is ``Enhance this underwater photo: correct the colour cast and improve clarity and contrast, without changing its content.'' (20 steps).
We measure noise with Immerk{\ae}r's estimator~\cite{immerkaer1996fast} and color change as the mean CIE76 $\Delta E^*_{ab}$ to step 1.
As in Banana100, each score is min-max normalized, and a score \emph{detects} the drift when its mean change from step 1 is negative at every checkpoint (steps 5, 10 and 20, and 50 on the replica chains).
Step 1, not the input, serves as the baseline, because the input has a different size.

\section{Agreement with human judgments}
\label{sec:agree}

\begin{figure}[t]
\centering
\includegraphics[width=\columnwidth]{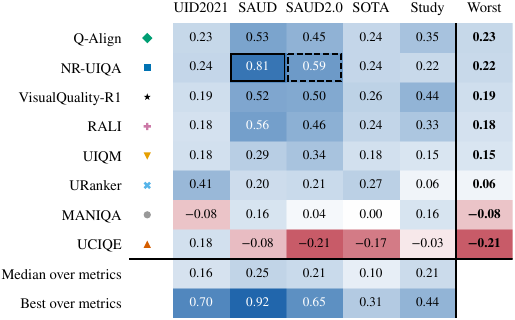}
\caption{\textbf{Agreement with people is low on every reference, and no metric performs well on all five.} Within-scene $\kappa$ (\cref{sec:kappa}) of the eight metrics of \cref{fig:grid}, sorted by their worst value over the five references (last column). Blue: agreement, red: the metric prefers the image people liked less. Solid frame: the metric's own training set. Dashed frame: a set from the same lab. Bottom rows: median and best value over all \nnMetrics metrics per reference (\nnAllFive have all five values). The best values on UID2021~\cite{hou2023uid2021} and SAUD~\cite{jiang2022underwater} come from metrics evaluated on their own training data. All metrics with their 95\% intervals are listed in \supp.}
\label{fig:kappa}
\end{figure}

\subsection{Agreement is low on all references}
\label{sec:agree_all}

No metric reaches $\kappa=$ \ninterRaterMed, the median agreement between two raters in our study, on all five references. This level comes from single raters on raw-versus-edit pairs, so for the opinion-score sets it is a reference point rather than a target. The best worst-case $\kappa$ is \nkWorstBest for \nkWorstBestName, followed by \nkWorstSecondName at \nkWorstSecond, with overlapping intervals (\cref{fig:kappa}).
On our study alone, \nkBestNameStudy reaches \nkBestStudy, above the agreement between two raters, so the gap lies in consistency across references. This $\kappa$ skips its ties. If a tie counts as half a correct choice, \nkBestNameStudy falls to rank \nvqAccTieRank with accuracy \nvqAccTie, and \naccTieBestName is best with \naccTieBest.
Much of this agreement comes from which edits a metric prefers. Choosing the people's majority answer for each edit gives $\kappa=$ \nweOracle, and replacing each metric's choices by its majority choice per edit changes the median $\kappa$ from \nweMedK to \nweMedEdit. With chance computed within each edit, two raters reach a median $\kappa$ of \nweRaterMed, \nweBestName \nweBest and \nweSecondName \nweSecond (\cref{tab:within_edit} in \supp).
The median metric reaches $\kappa=$ \nkMedUidA on UID2021, \nkMedSaud on SAUD, \nkMedSaudB on SAUD2.0~\cite{jiang2024toward}, \nkMedSota on SOTA~\cite{wang2022generation} and \nkMedStudy in our study.
High values occur only on single references: \nkBestNameSaud reaches \nkBestSaud on SAUD and \nkBestNameUidA \nkBestUidA on UID2021, both on their own training data.
On SOTA, the largest set, \nnSotaLowHalf of \nnSota metrics remain below 0.2.

The two standard underwater metrics are among the weakest.
UCIQE~\cite{yang2015underwater} has $\kappa$ \nkUciqeUidA on UID2021 and \nkUciqeSaud, \nkUciqeSaudB and \nkUciqeSota on SAUD, SAUD2.0 and SOTA, where its entire 95\% interval lies below zero on \nnUciqeBelowChance sets.
UIQM~\cite{panetta2015human} ranges between \nkUiqmUidA and \nkUiqmSaudB on the four opinion-score sets.

\paragraph{Ranking methods.}
When methods are ranked by mean within-scene position, UCIQE orders the enhancement methods with Kendall $\tau$ of \ntauUciqeUidA, \ntauUciqeSaud, \ntauUciqeSaudB and \ntauUciqeSota against the human order on the four sets, negative on \nnUciqeTauNeg of them. With 10 to 16 methods per set, these $\tau$ values indicate weak agreement rather than a reversed order.
UIQM gives \ntauUiqmUidA to \ntauUiqmSota.
On UID2021, only \nnTopHitUidA of \nnRankUidA metrics identify the method people liked best.
gMAD~\cite{ma2016group} and BAPPS-style 2AFC~\cite{zhang2018unreasonable} results are in \supp.

\subsection{Generalization of learned underwater metrics}
\label{sec:learned}

\begin{table}[t]
\centering\footnotesize\setlength{\tabcolsep}{3.5pt}
\resizebox{\columnwidth}{!}{\begin{tabular}{@{}lccccc@{}}
\toprule
Metric (trained on) & UID2021~\cite{hou2023uid2021} & SAUD~\cite{jiang2022underwater} & SAUD2.0~\cite{jiang2024toward} & SOTA~\cite{wang2022generation} & Study \\
\midrule
URanker~\cite{guo2023underwater} & \textbf{0.41} & 0.20 & 0.21 & \underline{0.27} & 0.06 \\
NUIQ~\cite{jiang2022underwater} (SAUD) & 0.22 & \cellcolor{insample}0.60 & \cellcolor{samelab}0.58 & 0.21 & 0.18 \\
UIQAVSI~\cite{chen2025underwater} (SAUD) & 0.18 & \cellcolor{insample}0.92 & \cellcolor{samelab}0.61 & 0.17 & 0.19 \\
NR-UIQA~\cite{yi2024no} (SAUD) & 0.24 & \cellcolor{insample}0.81 & \cellcolor{samelab}0.59 & 0.24 & 0.22 \\
Twice-Mixing~\cite{fu2022twice} & \underline{0.33} & 0.38 & 0.38 & 0.19 & 0.30 \\
QUIQ~\cite{karimi2025quiq} (UID2021) & \cellcolor{insample}0.70 & 0.25 & 0.23 & 0.11 & 0.24 \\
NMFC~\cite{chen2023underwater} (SAUD) & 0.31 & \cellcolor{insample}0.76 & \cellcolor{samelab}0.65 & \textbf{0.30} & 0.01 \\
\midrule
Q-Align~\cite{wu2023q} & 0.23 & \underline{0.53} & 0.45 & 0.24 & \underline{0.35} \\
VisualQuality-R1~\cite{wu2026visualquality} & 0.19 & 0.52 & \textbf{0.50} & 0.26 & \textbf{0.44} \\
RALI~\cite{zhao2026reasoning} & 0.18 & \textbf{0.56} & \underline{0.46} & 0.24 & 0.33 \\
\bottomrule
\end{tabular}
}
\caption{\textbf{Learned underwater metrics agree with people mainly on data from their own lab.} Within-scene $\kappa$. Dark cells: the metric's training set. Light cells: SAUD2.0, from the same lab as SAUD. Bold and underline mark the best and second best among the other cells of each column. Below the line: metrics not trained on underwater data. Twice-Mixing's~\cite{fu2022twice} training set is not among our references. URanker~\cite{guo2023underwater} was trained on enhancements of UIEB~\cite{li2019underwater} photos, including those of our study (\cref{sec:metrics}).}
\label{tab:learned}
\end{table}

Learned underwater metrics agree with people much more on their training set and same-lab data than elsewhere.
The five learned metrics with a known training set reach $\kappa$ \nkInMin to \nkInMax on it, \nkLabMin to \nkLabMax on SAUD2.0 from the same lab, and at most \nkOtherMax on data from other labs (\cref{tab:learned}).
Shared photos do not explain this drop.
Removing scenes whose photo also appears in the training set changes $\kappa$ by at most \nkDisjointDropMax.
Q-Align~\cite{wu2023q}, VisualQuality-R1~\cite{wu2026visualquality} and RALI~\cite{zhao2026reasoning}, which were never trained on underwater opinion scores, outperform every learned underwater metric outside its own lab on SAUD, SAUD2.0 and our study, but not on UID2021 and SOTA.

A random forest trained on \nfusNFeats metric scores from one set and tested on unseen photos of another set reaches at most $\kappa=$ \nfusOtherMax across labs (\nfusOtherMaxPair) and \nfusSaudBtoSaud from SAUD2.0 to SAUD (\supp).
The forest outperforms the best single metric in \nfusBeats of \nfusPairs set pairs, yet the best cross-lab forest remains below every in-sample value in \cref{tab:learned}.

\section{Manipulating metric scores}
\label{sec:gaming}

\begin{table}[t]
\centering
\resizebox{\columnwidth}{!}{\begin{tabular}{@{}lccccccccc@{}}
\toprule
 & \multicolumn{3}{c}{S+S} & \multicolumn{3}{c}{S+S+C} & \multicolumn{3}{c}{CLAHE} \\
\cmidrule(lr){2-4}\cmidrule(lr){5-7}\cmidrule(lr){8-10}
Metric & UID & SAUD & SOTA & UID & SAUD & SOTA & UID & SAUD & SOTA \\
\midrule
UIQM~\cite{panetta2015human} & 13 & 20 & 10 & 17 & 10 & 21 & 33 & 29 & 35 \\
UCIQE~\cite{yang2015underwater} & 42 & \textbf{61} & \textbf{81} & \textbf{60} & \textbf{86} & \textbf{95} & 33 & \textbf{59} & \textbf{76} \\
CCF~\cite{wang2018imaging} & 30 & \textbf{72} & \textbf{80} & \textbf{70} & \textbf{98} & \textbf{100} & \textbf{60} & \textbf{94} & \textbf{100} \\
FDUM~\cite{yang2021reference} & 8 & 47 & \textbf{78} & \textbf{70} & \textbf{93} & \textbf{99} & 32 & 47 & \textbf{88} \\
\midrule
URanker~\cite{guo2023underwater} & 2 & 1 & 4 & 32 & \textbf{51} & \textbf{64} & 33 & \textbf{56} & \textbf{72} \\
NR-UIQA~\cite{yi2024no} & 18 & 14 & 38 & 35 & 18 & 38 & 43 & 19 & 48 \\
Twice-Mixing~\cite{fu2022twice} & 3 & 4 & 20 & 37 & 26 & 48 & \textbf{62} & \textbf{69} & \textbf{89} \\
\midrule
NIQE~\cite{mittal2012making} & 45 & 45 & 43 & 10 & 28 & \textbf{50} & \textbf{60} & \textbf{73} & \textbf{76} \\
BRISQUE~\cite{mittal2012no} & \textbf{60} & 36 & 44 & 3 & 9 & 27 & 23 & 23 & 31 \\
MUSIQ~\cite{ke2021musiq} & \textbf{53} & 48 & \textbf{69} & \textbf{73} & 45 & \textbf{60} & \textbf{70} & 36 & \textbf{55} \\
CLIP-IQA+~\cite{wang2023exploring} & 42 & 16 & 36 & 22 & 6 & 14 & \textbf{63} & 32 & 44 \\
\midrule
Q-Align~\cite{wu2023q} & 20 & 9 & 27 & 8 & 8 & 8 & 42 & 15 & 26 \\
VisualQuality-R1~\cite{wu2026visualquality} & 20 & 9 & 26 & 8 & 4 & 6 & 18 & 11 & 26 \\
RALI~\cite{zhao2026reasoning} & \textbf{53} & 31 & 50 & 42 & 32 & 36 & \textbf{58} & 12 & 32 \\
\midrule
\multicolumn{10}{@{}l}{Metrics (of 71) where some edit beats the best in over half the scenes: UID2021 54, SAUD 42, SOTA 53} \\
\bottomrule
\end{tabular}
}
\caption{\textbf{A simple edit of the raw photo often outscores the enhancement people liked best.} Share of scenes (\%) in which the edit scores higher, bold above 50\%. S+S: saturation and unsharp masking. S+S+C: S+S followed by CLAHE. UID: UID2021 (\nnEditScenesUidA scenes), SAUD (\nnEditScenesSaud), SOTA (\nnEditScenesSota). All \nnEditMetrics metrics and six edits are listed in \supp.}
\label{tab:edits}
\end{table}

\subsection{Simple edits outscore the best enhancement}
For \nnEditHalfUidA of \nnEditMetrics metrics on UID2021~\cite{hou2023uid2021}, \nnEditHalfSaud on SAUD~\cite{jiang2022underwater} and \nnEditHalfSota on SOTA~\cite{wang2022generation}, at least one of our six edits outscores the enhancement people liked best in more than half of the scenes (\cref{tab:edits}).
UCIQE~\cite{yang2015underwater} prefers S+S+C to the human choice in \nuciqeSSCUidA, \nuciqeSSCSaud and \nuciqeSSCSota of scenes, and CCF~\cite{wang2018imaging} and FDUM~\cite{yang2021reference} rank the edits higher even more often.
Since no person compared these edits with the best enhancement, a moderate CLAHE~\cite{zuiderveld1994contrast} or S+S+C edit may sometimes look better.
Gaussian noise is not an enhancement, yet it outscores the best enhancement in up to \nnoiseWinUciqe of scenes under UCIQE, \nnoiseWinFdum under FDUM and \nnoiseWinQuiq under QUIQ~\cite{karimi2025quiq}. People did not compare noise with the best enhancement either, but in our study they reject strong noise even against the raw photo.
Only \nnEditSafe metrics stay at or below half on every set and edit: \neditSafeNames.

\begin{figure}[t]
\centering
\includegraphics[width=\columnwidth]{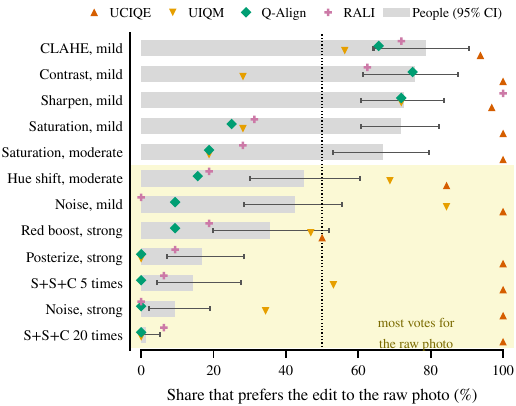}
\caption{\textbf{UCIQE prefers almost every edit, including those people reject.} One row per edit of our study, sorted by human preference. Gray bar: share of the \nnVotesPerCond votes that preferred the edit to the raw photo, with a 95\% interval from resampling raters and photos together. Markers: share of the \nnStudyImg photos for which the metric scores the edit above the raw photo. Yellow rows: edits that most people reject. Bars and markers are aggregate rates. Agreement on individual pairs is the study $\kappa$ (\cref{tab:kappa_all} in \supp). Per-edit values are listed in \supp.}
\label{fig:study}
\end{figure}

\subsection{People reject edits that metrics reward}
Our study confirms that the edits a metric rewards are not always the ones people prefer (\cref{fig:study}).
People prefer mild CLAHE, contrast, sharpening and saturation, as well as moderate saturation, to the raw photo, and reject \nnRejected of the \nnConditions edits, \nnRejectedSure of them with the entire 95\% interval below half (\nnRejectedPhoto with equal weight per photo, see \supp).
On these \nnRejected edits, UCIQE still prefers the edited photo in \nuciqeOnRejected of pairs, and UIQM~\cite{panetta2015human} in \nuiqmOnRejected.
After the saturate, sharpen and CLAHE chain is applied 20 times, \nhumanChainTwenty of votes prefer the result, whereas UCIQE prefers it for \nuciqeChainTwenty of photos.
Strong noise receives \nhumanNoiseStrong of votes and \nuciqeNoiseStrong of UCIQE's choices.
In contrast, Q-Align~\cite{wu2023q} and RALI~\cite{zhao2026reasoning} track people more closely and prefer few of the rejected edits.

\begin{table}[t]
\centering\footnotesize
\begin{tabular}{@{}lccccc@{}}
\toprule
 & \multicolumn{3}{c}{Clean raw photo beats best (\%)} & \multicolumn{2}{c}{Attacked, $\epsilon$ (\%)} \\
\cmidrule(lr){2-4}\cmidrule(lr){5-6}
Attacked metric & UID2021 & SAUD & SOTA & 2/255 & 8/255 \\
\midrule
MUSIQ~\cite{ke2021musiq} & 28 & 22 & 50 & 100 & 100 \\
LIQE~\cite{zhang2023blind} & 8 & 2 & 16 & 100 & 100 \\
TOPIQ-NR~\cite{chen2024topiq} & 54 & 32 & 50 & 100 & 100 \\
CLIP-IQA+~\cite{wang2023exploring} & 66 & 42 & 64 & 100 & 100 \\
RALI~\cite{zhao2026reasoning} & 40 & 20 & 44 & 100 & 100 \\
\bottomrule
\end{tabular}

\caption{\textbf{Ten gradient steps make each attacked metric prefer the raw photo.} Share of the \nnAttScenes scenes per set in which the raw photo scores above the people's best enhancement, before and after a PGD~\cite{madry2017towards} attack on that metric. Mean PSNR of the attacked photo: \npsnrMeanTwo\,dB at $\epsilon=2/255$ and \npsnrMeanEight\,dB at $8/255$.}
\label{tab:attack}
\end{table}

\subsection{Small gradient steps succeed against every attacked metric}
With $\epsilon=2/255$, ten PGD steps raise the raw photo above the best human-rated enhancement in \nattWin of the \nnAttScenesAll scenes for all \nnAttMetrics attacked metrics (\cref{tab:attack}), while the photo retains a PSNR of at least \npsnrMinTwo\,dB.
The attacks partly transfer.
A photo attacked through RALI at $8/255$ also outscores the best enhancement under MUSIQ~\cite{ke2021musiq} in \nraliToMusiq of scenes, compared with \ncleanMusiq before the attack (\supp).

\section{Sensitivity to generative drift}
\label{sec:drift}

\begin{figure*}[t]
\centering
\includegraphics[width=\textwidth]{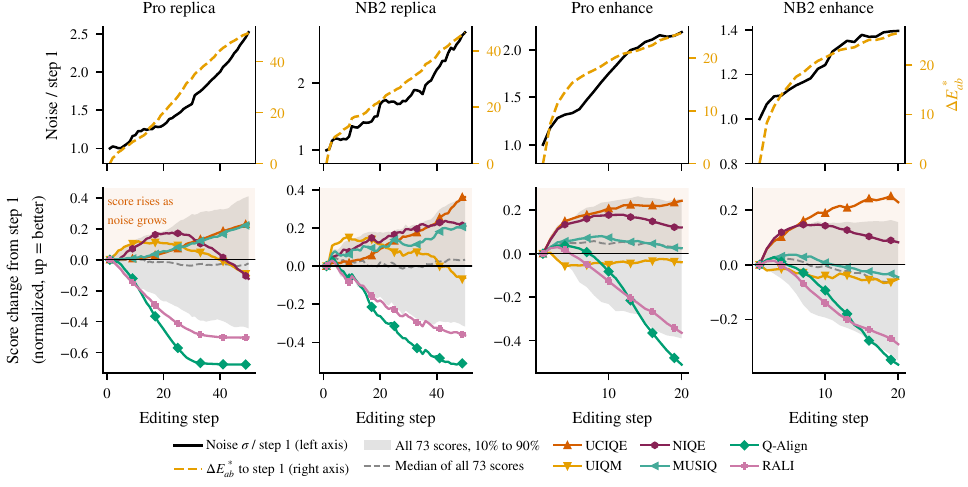}
\caption{\textbf{Noise and color change grow along every chain, while UCIQE rises on every chain and NIQE and MUSIQ rise on most.} Top: mean noise $\sigma$ relative to step 1 (black, left axis) and mean $\Delta E^*_{ab}$ to step 1 (orange, right axis). Bottom: each score min-max normalized per photo, minus its step 1 value, so up means that the metric rates the photo better than at step 1. Gray band and dashed line: 10th to 90th percentile and median of all \nnDriftScores scores. Shaded area above zero: the score rises although noise and color change grow. Means over the \nnDriftImg photos. The last frames are shown in \cref{fig:grid}, and crops along each chain in \cref{fig:drift_img} of \supp.}
\label{fig:drift}
\end{figure*}

\begin{table}[t]
\centering
\resizebox{\columnwidth}{!}{\begin{tabular}{@{}lcccc@{}}
\toprule
 & Pro replica & NB2 replica & Pro enhance & NB2 enhance \\
\midrule
Steps & 50 & 50 & 20 & 20 \\
Noise $\sigma$ (8-bit levels), first to last step & 1.0 to 2.6 & 1.0 to 2.8 & 2.0 to 4.3 & 2.3 to 3.2 \\
Images whose noise grows & 20/20 & 19/20 & 20/20 & 17/20 \\
Color change $\Delta E^*_{ab}$, last to first step & 51.2 & 46.3 & 24.7 & 26.6 \\
Mean $a^*$, first to last step & $-$15.9 to 15.7 & $-$15.3 to 8.3 & $-$12.0 to 0.4 & $-$10.3 to 0.8 \\
\midrule
Scores that detect drift (of 73) & 22 & 11 & 19 & 26 \\
\midrule
UIQM~\cite{panetta2015human} & \xmark $-$0.11 & \xmark $-$0.07 & \cmark $-$0.05 & \cmark $-$0.06 \\
UCIQE~\cite{yang2015underwater} & \xmark 0.24 & \xmark 0.36 & \xmark 0.26 & \xmark 0.25 \\
UIConM~\cite{panetta2015human} & \xmark $-$0.54 & \xmark $-$0.50 & \cmark $-$0.30 & \cmark $-$0.23 \\
NIQE~\cite{mittal2012making} & \xmark $-$0.14 & \xmark 0.24 & \xmark 0.15 & \xmark 0.12 \\
BRISQUE~\cite{mittal2012no} & \xmark $-$0.20 & \xmark 0.13 & \xmark 0.36 & \xmark 0.27 \\
MUSIQ~\cite{ke2021musiq} & \xmark 0.23 & \xmark 0.21 & \xmark 0.04 & \xmark $-$0.07 \\
URanker~\cite{guo2023underwater} & \xmark $-$0.06 & \xmark $-$0.15 & \xmark 0.01 & \xmark $-$0.05 \\
Q-Align~\cite{wu2023q} & \cmark $-$0.68 & \xmark $-$0.51 & \xmark $-$0.51 & \cmark $-$0.41 \\
VisualQuality-R1~\cite{wu2026visualquality} & \cmark $-$0.69 & \xmark $-$0.56 & \xmark $-$0.62 & \xmark $-$0.40 \\
RALI~\cite{zhao2026reasoning} & \cmark $-$0.50 & \cmark $-$0.36 & \cmark $-$0.44 & \cmark $-$0.38 \\
\bottomrule
\end{tabular}
}
\caption{\textbf{Few scores detect the drift, and UCIQE rises on every chain.} Top: image measures, means over \nnDriftImg photos. Bottom: mean normalized score change from step 1 to the last step, \cmark\ when the score falls below step 1 at every checkpoint (Banana100's~\cite{tang2026banana100} rule). All \nnDriftScores scores and per-photo pass rates are listed in \supp.}
\label{tab:drift}
\end{table}

Under the strict rule of Banana100~\cite{tang2026banana100}, most metrics do not detect the drift.
Every chain visibly changes the image (\cref{fig:grid} and \cref{fig:drift_img} of \supp).
Noise increases in \nnSigmaUp photo chains, mean noise $\sigma$ rises by $\times$\nsigmaRatioMin to $\times$\nsigmaRatioMax from step 1 to the last step, and the color shifts by $\Delta E^*_{ab}$ \ndEMin to \ndEMax (\cref{tab:drift}).
The replica chains also turn the green cast red (mean $a^*$ becomes positive).
These measures do not prove that people perceive a worse image. On the enhance chains, part of the color change is the requested correction, and the NB2 replica chain redraws the scene into a clean but different image (\cref{fig:drift_img}). The Pro replica chain, which should change nothing yet becomes noisier, is therefore the clearest test.

However, only \nnDetectProRep, \nnDetectNbRep, \nnDetectProEnh and \nnDetectNbEnh of \nnDriftScores scores detect the drift on the Pro replica, NB2 replica, Pro enhance and NB2 enhance chains.
Only \nnAllFour detect it on all four: \nallFourNames.
Only RALI~\cite{zhao2026reasoning} among these is a default checkpoint.
This rule is strict. A looser rule that checks only the last step credits \nnLastDownProRep scores on the Pro replica chain and \nnLastDownAll on all four chains, including UIQM~\cite{panetta2015human}, Q-Align~\cite{wu2023q} and VisualQuality-R1~\cite{wu2026visualquality}, but still not UCIQE~\cite{yang2015underwater}.
UCIQE rises on all four chains, by \nuciqeDProRep, \nuciqeDNbRep, \nuciqeDProEnh and \nuciqeDNbEnh in normalized units, and decreases for at most \nuciqePerImgProEnh of photos on any chain.
Among the hand-crafted underwater scores, only \nhandDetectAny detect any chain, and only the enhance chains.
The vision-language scores decrease on average on every chain (\cref{fig:drift}), but not at every checkpoint.

Per photo, the best score decreases on \nperImgBestProRep of Pro replica photos (\nperImgBestNameProRep) but only on \nperImgBestNbRep of NB2 replica photos (\nperImgBestNameNbRep).

\section{Causes of metric failure}
\label{sec:why}

\begin{table}[t]
\centering\footnotesize\setlength{\tabcolsep}{3pt}
\resizebox{\columnwidth}{!}{\begin{tabular}{@{}lccccc@{}}
\toprule
 & UID2021~\cite{hou2023uid2021} & SAUD~\cite{jiang2022underwater} & SAUD2.0~\cite{jiang2024toward} & SOTA~\cite{wang2022generation} & Study \\
\midrule
NaN share, MATLAB code as released & 32\% & 38\% & 34\% & 27\% & n/a \\
$\kappa$, our NumPy port & 0.18 & $-$0.08 & $-$0.21 & $-$0.17 & $-$0.03 \\
$\kappa$, MATLAB code, NaN patched & 0.34 & 0.00 & $-$0.15 & $-$0.09 & 0.02 \\
\bottomrule
\end{tabular}
}
\caption{\textbf{Two UCIQE implementations give different answers.} The public MATLAB code returns NaN for every image with a pure-black pixel. With these NaNs replaced, its $\kappa$ differs from that of our NumPy port by up to \nkUciqeImplDiff.}
\label{tab:uciqe_impl}
\end{table}

We examine three possible causes and label the untested parts as hypotheses.

\paragraph{The UCIQE formula rewards the edits.}
UCIQE is a weighted sum of the chroma spread, the luminance contrast and the mean saturation~\cite{yang2015underwater}.
Saturation increases the first and third terms, unsharp masking and CLAHE~\cite{zuiderveld1994contrast} increase the second, and noise increases the first two.
This is consistent with UCIQE's preference for S+S+C (\cref{tab:edits}), strong noise (\cref{fig:study}) and the noisy, red-shifted drift frames (\cref{tab:drift}).
We hypothesize that any score built from such global statistics, without a model of how the scene should look, can be raised in this way.

\paragraph{The public code does not always return a score.}
The public MATLAB UCIQE code~\cite{wong2020uciqecode}, run in GNU Octave~\cite{octave2022gnu}, returns NaN whenever an image contains a pure-black pixel, which occurs for \nnanMin to \nnanMax of the opinion-score images.
Even on images for which both return a score, our NumPy port and the MATLAB code rank images differently (Spearman correlation \nimplRhoMin to \nimplRhoMax per set).
With the NaN patched, the two implementations differ in $\kappa$ (\cref{tab:uciqe_impl}), and the patched MATLAB UCIQE rises even more along the drift chains (\nuciqeMatDProRep on Pro replica).
Both versions divide chroma by $L^*$, which is unstable for near-black pixels. For example, \nuciqeDarkShare of the pixels in the last Pro replica frame of UIEB~\cite{li2019underwater} photo \nuciqeDarkId have $L^*<1$, which produces the port's value of \nuciqeDarkVal in \cref{fig:grid}.
The qualitative UCIQE findings hold with the patched MATLAB code, although the magnitudes differ: S+S+C outscores the best enhancement in \nuciqeMatSSCMin to \nuciqeMatSSCMax of scenes (\nuciqeSSCUidA to \nuciqeSSCMax for our port), its $\kappa$ in our study is \nkUciqeMatStudy, and it rises on all four drift chains, by \nuciqeMatDNbEnh to \nuciqeMatDProRep.
Only its $\kappa$ on UID2021~\cite{hou2023uid2021} (\nkUciqeMatUidA) is clearly higher than that of the port.

\paragraph{Learned metrics change their choice under flips and resizing.}
Horizontally flipping both images of a pair reverses a metric's preference in a median of \nflipMedHflip of within-scene pairs (up to \nflipMaxHflip), and halving the image size does so in \nflipMedHalf (up to \nflipMaxHalf), whereas UCIQE and UIQM~\cite{panetta2015human}, being global statistics, are unaffected by flips (\supp).
A flip does not change quality.
We hypothesize that part of the low agreement of learned metrics stems from this sensitivity rather than from an incorrect notion of quality.

\section{Recommendations and conclusion}
\label{sec:conclusion}

\paragraph{Five recommendations.}
Our results suggest five checks for any paper that ranks underwater enhancement methods with an NR-IQA metric.
(1)~Report within-scene agreement with people on at least two references from different labs, and report the lower value.
(2)~Do not rank methods by UCIQE~\cite{yang2015underwater} alone, since its $\kappa$ is near or below zero on three of the four opinion-score sets with either implementation (\cref{tab:uciqe_impl}).
(3)~Include the raw photo and a simple S+S+C edit as baselines. A method that cannot outperform them under the chosen metric has not shown a clear gain, and a metric under which they often win should be checked against people before it is used to rank methods.
(4)~Report the implementation and version of every metric.
(5)~When the enhancement method is generative, measure noise and color change directly, because most metrics do not decrease as these grow.

\paragraph{Limitations.}
Our study is small: \nnRatersAll raters, all from our university, \nnStudyImg photos and \nnConditions edits, viewed on the raters' own screens, and the raters agree with each other only moderately (median $\kappa=$ \ninterRaterMed). Blocks had \nblockSizes raters, so some photos have few votes. Pairs whose opinion scores differ by less than their rating noise are still counted, which can lower $\kappa$.
Its references are raw UIEB~\cite{li2019underwater} photos rather than enhancement outputs, and no person compared a simple edit with the best enhancement or a late drift frame with an early one.
The drift test uses \nnDriftImg photos, two editors and two prompts, and the outputs of the editors may change between model versions.
Simple edits and attacks were applied to at most \nnEditScenesSota and \nnAttScenes scenes per set rather than to every scene, and the attacks target only \nnAttMetrics general-purpose metrics, not the learned underwater metrics.
We could not run every published underwater metric, and two learned metrics have training sets outside our references.
The causes in \cref{sec:why} beyond UCIQE's formula and code are untested hypotheses.

\paragraph{Societal impact.}
Underwater images support marine surveys and inspection, and a metric that rewards noise or color shifts can promote enhancement methods that alter what a scientist or inspector sees.
We report our score manipulation tests so that evaluations can guard against them.
All photos come from public research datasets used under their licenses.
Raters agreed to take part on the study page before starting, could stop at any time by closing it, and were recorded without names: choices, response times, screen size, an age band and self-rated experience.

\paragraph{Release.}
The code is available at \codeurl and \dsname at \dataurl, with the study choices, all scores and $\kappa$ tables and the images we generated from UIEB photos.
For the opinion-score sets and UIEB, which remain with their authors, we release image lists and scripts that rebuild our edits and attacks.

\paragraph{Conclusion.}
Across \nnMetrics metrics and five human references, no metric reaches the median agreement between two of our raters on every reference.
Learned underwater metrics agree with people mainly on data from their own lab, simple edits and small gradient steps outscore the enhancement people liked best, and few metrics detect the noise and color shifts of repeated generative edits.
A higher underwater quality score alone is not evidence of a better image.

\paragraph{Acknowledgments.}
We thank the \nnRatersAll raters who took part in our study and the Artificial Intelligence and Machine Learning Society (AIMS) of Delhi Technological University.

{
    \small
    \bibliographystyle{ieeenat_fullname}
    \bibliography{scholar,references}
}

\renewcommand{\theHtable}{S\arabic{table}}\renewcommand{\theHfigure}{S\arabic{figure}}
\appendix
\renewcommand{\thetable}{S\arabic{table}}
\renewcommand{\thefigure}{S\arabic{figure}}
\setcounter{table}{0}
\setcounter{figure}{0}
\renewcommand{\topfraction}{0.95}\renewcommand{\dbltopfraction}{0.95}\renewcommand{\textfraction}{0.05}
\renewcommand{\floatpagefraction}{0.6}\renewcommand{\dblfloatpagefraction}{0.6}
\setcounter{topnumber}{4}\setcounter{dbltopnumber}{4}\setcounter{totalnumber}{6}
\twocolumn[{\begin{center}{\Large\bf Appendix}\end{center}\vspace{0.5em}
\centering
\includegraphics[width=\textwidth]{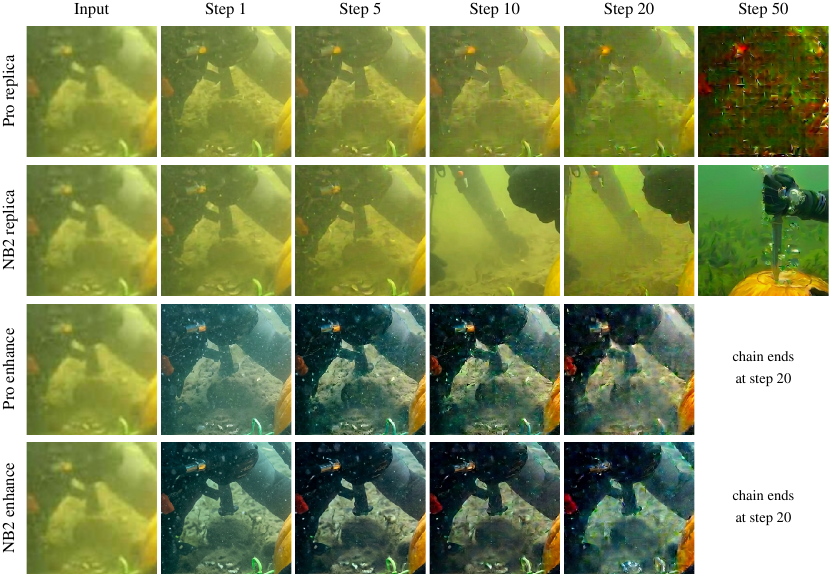}
\captionof{figure}{\textbf{Repeated generative edits add noise, shift color and can replace the content.} Center crops ($300\times300$\,px) of UIEB photo \ndriftCropId, the first row block of \cref{fig:grid}, along the four chains. Every frame is resized to the Pro replica output size of this photo ($\ndriftFrame$\,px) before cropping. The input is \ndriftInW\,px wide. The NB2 replica chain redraws the scene from step 10 on, so its later crops no longer match the input.}
\label{fig:drift_img}
\vspace{1em}}]
\phantomsection\label{sec:appendix}

\begin{table*}[t]
\centering\small
\begin{tabular}{@{}lcccccc@{}}
\toprule
 & People & \multicolumn{5}{c}{Metric prefers the edit (\%)} \\
\cmidrule(lr){3-7}
Edit of the raw photo & prefer edit (\%) & UCIQE~\cite{yang2015underwater} & UIQM~\cite{panetta2015human} & NIQE~\cite{mittal2012making} & Q-Align~\cite{wu2023q} & RALI~\cite{zhao2026reasoning} \\
\midrule
CLAHE, mild & 79 {\scriptsize [64, 91]} & 94 & 56 & 72 & 66 & 72 \\
Contrast, mild & 76 {\scriptsize [61, 87]} & 100 & 28 & 72 & 75 & 62 \\
Sharpen, mild & 72 {\scriptsize [61, 84]} & 97 & 72 & 69 & 72 & 100 \\
Saturation, mild & 72 {\scriptsize [61, 82]} & 100 & 28 & 72 & 25 & 31 \\
Saturation, moderate & 67 {\scriptsize [53, 79]} & 100 & 19 & 75 & 19 & 28 \\
Hue shift, moderate & 45 {\scriptsize [30, 60]} & 84 & 69 & 72 & 16 & 19 \\
Noise, mild & 42 {\scriptsize [28, 55]} & 100 & 84 & 0 & 9 & 0 \\
Red boost, strong & 36 {\scriptsize [20, 52]} & 50 & 47 & 75 & 9 & 19 \\
Posterize, strong & 17 {\scriptsize [7, 28]} & 100 & 0 & 0 & 0 & 9 \\
S+S+C chain, 5 steps & 14 {\scriptsize [4, 28]} & 100 & 53 & 3 & 0 & 6 \\
Noise, strong & 9 {\scriptsize [2, 19]} & 100 & 34 & 0 & 0 & 0 \\
S+S+C chain, 20 steps & 1 {\scriptsize [0, 5]} & 100 & 0 & 0 & 0 & 6 \\
\bottomrule
\end{tabular}

\caption{\textbf{People and metrics on the \nnConditions edits of our study.} People: share of the \nnVotesPerCond votes that prefer the edit, with 95\% interval from resampling raters and photos together. Metrics: share of the \nnStudyImg photos where the metric scores the edit higher.}
\label{tab:study_supp}
\end{table*}

\begin{table}[t]
\centering\small\setlength{\tabcolsep}{3pt}
\resizebox{\columnwidth}{!}{\begin{tabular}{@{}lccc@{}}
\toprule
 & $\kappa$ & Edit preference only & Within edit [95\% interval] \\
\midrule
Two raters (median) & 0.41 & n/a & 0.13 \\
People's majority per edit & 0.49 & 0.49 & 0.00 \\
\midrule
LIQE~\cite{zhang2023blind} & 0.42 & 0.49 & 0.15 {\scriptsize [0.05, 0.24]} \\
Q-Align~\cite{wu2023q} & 0.35 & 0.33 & 0.12 {\scriptsize [0.05, 0.18]} \\
VisualQuality-R1~\cite{wu2026visualquality} & 0.44 & 0.50 & 0.10 {\scriptsize [0.02, 0.18]} \\
LAION-Aes~\cite{schuhmann2024christophschuhmann} & 0.40 & 0.45 & 0.09 {\scriptsize [0.01, 0.16]} \\
QUIQ~\cite{karimi2025quiq} & 0.24 & 0.06 & 0.08 {\scriptsize [0.02, 0.14]} \\
RALI~\cite{zhao2026reasoning} & 0.33 & 0.33 & 0.08 {\scriptsize [0.00, 0.14]} \\
MANIQA-KADID~\cite{yang2022maniqa} & 0.29 & 0.19 & 0.08 {\scriptsize [0.00, 0.16]} \\
MUSIQ-AVA~\cite{ke2021musiq} & 0.35 & 0.39 & 0.08 {\scriptsize [$-$0.01, 0.16]} \\
UIQM~\cite{panetta2015human} & 0.15 & 0.11 & 0.05 {\scriptsize [$-$0.03, 0.12]} \\
UCIQE~\cite{yang2015underwater} & $-$0.03 & 0.00 & $-$0.03 {\scriptsize [$-$0.07, 0.00]} \\
\bottomrule
\end{tabular}
}
\caption{\textbf{In our study, agreement within an edit is low for people and metrics alike.} $\kappa$: as in the main paper. Edit preference only: each metric's choice replaced by its majority choice for that edit, over the \nnStudyImg photos. Within edit: chance agreement computed inside each edit from its own two marginals and pooled over edits, so a rule that knows only the edit scores 0. Interval: raters and photos resampled together. The eight metrics with the highest within-edit $\kappa$ of \nweN, plus UIQM and UCIQE.}
\label{tab:within_edit}
\end{table}

\begin{table*}[t]
\centering\small
\begin{tabular}{@{}lllll@{}}
\toprule
Rank & UID2021~\cite{hou2023uid2021} & SAUD~\cite{jiang2022underwater} & SAUD2.0~\cite{jiang2024toward} & SOTA~\cite{wang2022generation} \\
\midrule
1 & QUIQ~\cite{karimi2025quiq} (1.00) & UIQAVSI~\cite{chen2025underwater} (1.00) & UIQAVSI~\cite{chen2025underwater} (0.99) & Q-Align~\cite{wu2023q} (0.99) \\
2 & URanker~\cite{guo2023underwater} (0.95) & NR-UIQA~\cite{yi2024no} (0.99) & NUIQ~\cite{jiang2022underwater} (0.99) & UNIQUE~\cite{zhang2021uncertainty} (0.98) \\
3 & Entropy (0.93) & NMFC~\cite{chen2023underwater} (0.99) & NMFC~\cite{chen2023underwater} (0.98) & RALI~\cite{zhao2026reasoning} (0.96) \\
4 & NMFC~\cite{chen2023underwater} (0.89) & RALI~\cite{zhao2026reasoning} (0.96) & RALI~\cite{zhao2026reasoning} (0.98) & VisualQuality-R1~\cite{wu2026visualquality} (0.95) \\
5 & FDUM~\cite{yang2021reference} (0.86) & NUIQ~\cite{jiang2022underwater} (0.96) & QualiCLIP~\cite{agnolucci2024quality} (0.98) & MANIQA-KADID~\cite{yang2022maniqa} (0.93) \\
\midrule
UCIQE & 27 of 73 (0.76) & 70 of 73 (0.49) & 73 of 73 (0.35) & 71 of 72 (0.11) \\
\bottomrule
\end{tabular}

\caption{\textbf{gMAD aggressiveness with the opinion scores as judge.} The five metrics best at finding image pairs where another metric fails, and the rank of UCIQE, per set. Value in brackets: mean win rate as attacker.}
\label{tab:gmad}
\end{table*}

\begin{table*}[t]
\centering\small
\begin{tabular}{@{}lcccc@{}}
\toprule
Share of within-scene pairs whose order flips & H. flip & 90$^\circ$ turn & Half size & 90\% crop \\
\midrule
Median over metrics & 7\% & 14\% & 27\% & 10\% \\
Maximum over metrics & 31\% & 39\% & 47\% & 35\% \\
UIQM~\cite{panetta2015human} & 0\% & 0\% & 17\% & 3\% \\
UCIQE~\cite{yang2015underwater} & 0\% & 0\% & 2\% & 3\% \\
MUSIQ~\cite{ke2021musiq} & 10\% & 22\% & 30\% & 12\% \\
Q-Align~\cite{wu2023q} & 6\% & 12\% & 33\% & 10\% \\
RALI~\cite{zhao2026reasoning} & 7\% & 18\% & 38\% & 11\% \\
\bottomrule
\end{tabular}

\caption{\textbf{Flips, turns, resizing and cropping reverse many within-scene choices.} Share of UID2021 pairs whose order changes when both images are flipped, turned, halved or cropped. Median and maximum over \nnRobust metrics.}
\label{tab:robust}
\end{table*}

\begin{table*}[t]
\centering\small
\setlength{\tabcolsep}{5pt}\begin{tabular}{@{}lccccccc@{}}
\toprule
Attacked $\downarrow$ / scored by $\rightarrow$ & MUSIQ~\cite{ke2021musiq} & LIQE~\cite{zhang2023blind} & TOPIQ-NR~\cite{chen2024topiq} & CLIP-IQA+~\cite{wang2023exploring} & RALI~\cite{zhao2026reasoning} & UIQM~\cite{panetta2015human} & UCIQE~\cite{yang2015underwater} \\
\midrule
Clean (no attack) & 33 & 9 & 45 & 57 & 35 & 17 & 27 \\
\midrule
MUSIQ~\cite{ke2021musiq} & 100 & 25 & 50 & 56 & 11 & 47 & 32 \\
LIQE~\cite{zhang2023blind} & 54 & 100 & 56 & 39 & 51 & 36 & 28 \\
TOPIQ-NR~\cite{chen2024topiq} & 32 & 33 & 100 & 82 & 13 & 47 & 31 \\
CLIP-IQA+~\cite{wang2023exploring} & 13 & 16 & 51 & 100 & 6 & 46 & 32 \\
RALI~\cite{zhao2026reasoning} & 61 & 45 & 55 & 55 & 100 & 33 & 30 \\
\bottomrule
\end{tabular}

\caption{\textbf{Gradient attacks partly transfer.} Share of scenes (\%) where the raw photo, attacked through the row metric at $\epsilon=8/255$, beats the people's best enhancement under the column metric. First row: no attack. \nnAttScenesAll scenes from UID2021, SAUD and SOTA.}
\label{tab:transfer}
\end{table*}

This material provides the setup details and full tables behind the main paper. Every number is generated by \texttt{scripts/make\_numbers.py} from the result files.
All metrics ran on one NVIDIA RTX 5070 Ti with 16\,GB. The \nnEditorImages Nano Banana images cost about US\$\neditorCost at Google's list prices for the logged tokens.

\section{Edits, prompts and study details}
\label{sec:a_setup}

\paragraph{Edit strengths.} %
Images are in $[0,1]$. Saturation multiplies HSV saturation by 1.3, 1.6 or 2.0 (mild, moderate, strong). Unsharp masking adds $a(x-G_2*x)$ with a Gaussian of $\sigma=2$ and $a=0.5$, 1.5 or 3.0. CLAHE works on the $L^*$ channel with $8\times8$ tiles and clip limit 2, 4 or 8. Contrast scales the deviation from the image mean by 1.3, 1.6 or 2.0. Gaussian noise has standard deviation 0.02, 0.05 or 0.10. The red gain multiplies the red channel by 1.2, 1.5 or 2.0. The hue shift is 15, 45 or 90 degrees, and posterization keeps 5, 4 or 3 bits. One step of the S+S+C chain of our study applies saturation $\times1.15$, unsharp masking with $a=0.5$ and CLAHE with clip limit 2.

\paragraph{Generative editor prompts.} Replica (Banana100's prompt): ``Produce an exact replica of the provided image, with no alterations.'' Enhance: ``Enhance this underwater photo: correct the colour cast and improve clarity and contrast, without changing its content.'' Every step is a new request that contains only the prompt and the previous output. \Cref{fig:drift_img} shows crops of one photo along the four chains.

\paragraph{Study.} \nnStudyImg UIEB photos were split into four blocks of eight. Each rater saw one block: 96 test pairs (8 photos $\times$ \nnConditions edits), 6 attention checks against a strongly blurred copy, 5 repeated pairs and 4 pairs of identical images, in random order and random side. Two practice pairs on other photos came first. Each block had its own link. The four blocks were seen by \nblockSizes raters, so a photo received \nnBlockSmall to \nnBlockBig votes per edit, and the vote shares in \cref{fig:study} weight photos in larger blocks more. With equal weight per photo, only the moderate hue shift crosses half, from \nhueVote to \nhuePhoto, so \nnRejectedPhoto edits are rejected. The \nnRejectedSure edits whose interval lies below half stay at or below \nsurePhotoMax. Raters gave the same answer on \nrepeatSame of the repeated pairs and picked the left image on \nidentLeft of the identical pairs. \Cref{tab:study_supp} gives the result per edit.

\section{Agreement and generalization}

\paragraph{gMAD and paired choices.}
In a gMAD-style competition~\cite{ma2016group} on the rated pairs, UCIQE ranks \ngmadUciqeRankSota of \ngmadNSota on SOTA in exposing failures of other metrics (all sets in \cref{tab:gmad}).
In BAPPS-style 2AFC scoring~\cite{zhang2018unreasonable} on our study, \ntwoafcAtCeiling metrics reach the rater-to-rater level of \ntwoafcCeiling (best \ntwoafcBestName, \ntwoafcBest), but the median metric scores \ntwoafcMedian and UCIQE \ntwoafcUciqe, below the 0.5 of a coin flip.

\paragraph{Study agreement within an edit.}
\Cref{tab:within_edit} splits the study $\kappa$ into the part that comes from which edits a metric prefers and the part that comes from its choices per photo within an edit. \nweAboveZero of the \nweN metrics have a within-edit interval above zero, and only \nweBestName reaches the median of two raters. UCIQE has \nweUciqe.

\begin{table}[!ht]
\centering\small\setlength{\tabcolsep}{3pt}
\resizebox{\columnwidth}{!}{\begin{tabular}{@{}llcc@{}}
\toprule
Trained on & Tested on & Forest $\kappa$ & Best single \\
\midrule
UID2021~\cite{hou2023uid2021} & SAUD~\cite{jiang2022underwater} & 0.34 {\scriptsize [0.30, 0.38]} & FDUM~\cite{yang2021reference} (0.08) \\
UID2021~\cite{hou2023uid2021} & SAUD2.0~\cite{jiang2024toward} & 0.49 {\scriptsize [0.47, 0.52]} & FDUM~\cite{yang2021reference} (0.05) \\
UID2021~\cite{hou2023uid2021} & SOTA~\cite{wang2022generation} & 0.17 {\scriptsize [0.15, 0.19]} & FDUM~\cite{yang2021reference} ($-$0.12) \\
SAUD~\cite{jiang2022underwater} & UID2021~\cite{hou2023uid2021} & 0.27 {\scriptsize [0.22, 0.31]} & RALI~\cite{zhao2026reasoning} (0.15) \\
SAUD~\cite{jiang2022underwater} & SAUD2.0~\cite{jiang2024toward} & 0.61 {\scriptsize [0.58, 0.63]} & RALI~\cite{zhao2026reasoning} (0.46) \\
SAUD~\cite{jiang2022underwater} & SOTA~\cite{wang2022generation} & 0.28 {\scriptsize [0.26, 0.30]} & RALI~\cite{zhao2026reasoning} (0.24) \\
SAUD2.0~\cite{jiang2024toward} & UID2021~\cite{hou2023uid2021} & 0.30 {\scriptsize [0.26, 0.34]} & NMFC~\cite{chen2023underwater} (0.30) \\
SAUD2.0~\cite{jiang2024toward} & SAUD~\cite{jiang2022underwater} & 0.63 {\scriptsize [0.59, 0.67]} & VisualQuality-R1~\cite{wu2026visualquality} (0.53) \\
SAUD2.0~\cite{jiang2024toward} & SOTA~\cite{wang2022generation} & 0.35 {\scriptsize [0.33, 0.36]} & NMFC~\cite{chen2023underwater} (0.30) \\
SOTA~\cite{wang2022generation} & UID2021~\cite{hou2023uid2021} & 0.25 {\scriptsize [0.20, 0.31]} & ARNIQA-LIVE~\cite{agnolucci2024arniqa} (0.04) \\
SOTA~\cite{wang2022generation} & SAUD~\cite{jiang2022underwater} & 0.39 {\scriptsize [0.35, 0.42]} & ARNIQA-LIVE~\cite{agnolucci2024arniqa} (0.40) \\
SOTA~\cite{wang2022generation} & SAUD2.0~\cite{jiang2024toward} & 0.58 {\scriptsize [0.55, 0.60]} & ARNIQA-LIVE~\cite{agnolucci2024arniqa} (0.31) \\
\bottomrule
\end{tabular}
}
\caption{\textbf{Random-forest fusion of \nfusNFeats scores across sets.} Trained on one set's opinion scores, tested within scene on the other set's photos not shared with the training set. Metrics trained on either set are left out. Best single: the metric with the highest $\kappa$ on the training set, scored on the test set. Brackets: 95\% scene-bootstrap interval.}
\label{tab:fusion}
\end{table}

\onecolumn
\section{Full tables}

{\footnotesize\setlength{\tabcolsep}{4pt}
\begin{longtable}{@{}lccccc@{}}
\caption{Within-scene Cohen's $\kappa$ of every metric with people, with 95\% scene-bootstrap intervals (study: rater and image cluster bootstrap). n/a: not run. Groups as in Sec.~3.2 of the main paper.}\label{tab:kappa_all}\\
\toprule
Metric & UID2021~\cite{hou2023uid2021} & SAUD~\cite{jiang2022underwater} & SAUD2.0~\cite{jiang2024toward} & SOTA~\cite{wang2022generation} & Our study \\
\midrule
\endfirsthead
\toprule
Metric & UID2021~\cite{hou2023uid2021} & SAUD~\cite{jiang2022underwater} & SAUD2.0~\cite{jiang2024toward} & SOTA~\cite{wang2022generation} & Our study \\
\midrule
\endhead
\multicolumn{6}{@{}l}{\emph{Hand-crafted underwater}} \\
CCF~\cite{wang2018imaging} & 0.29 {\scriptsize [0.25, 0.34]} & $-$0.08 {\scriptsize [$-$0.12, $-$0.04]} & $-$0.20 {\scriptsize [$-$0.22, $-$0.19]} & $-$0.09 {\scriptsize [$-$0.11, $-$0.08]} & $-$0.06 {\scriptsize [$-$0.13, 0.00]} \\
FDUM~\cite{yang2021reference} & 0.43 {\scriptsize [0.38, 0.48]} & 0.09 {\scriptsize [0.06, 0.13]} & 0.05 {\scriptsize [0.02, 0.08]} & $-$0.12 {\scriptsize [$-$0.13, $-$0.10]} & 0.09 {\scriptsize [0.03, 0.16]} \\
UCIQE~\cite{yang2015underwater} & 0.18 {\scriptsize [0.14, 0.23]} & $-$0.08 {\scriptsize [$-$0.12, $-$0.04]} & $-$0.21 {\scriptsize [$-$0.23, $-$0.19]} & $-$0.17 {\scriptsize [$-$0.19, $-$0.16]} & $-$0.03 {\scriptsize [$-$0.08, 0.02]} \\
UICM~\cite{panetta2015human} & 0.34 {\scriptsize [0.30, 0.38]} & 0.03 {\scriptsize [$-$0.01, 0.07]} & $-$0.01 {\scriptsize [$-$0.04, 0.02]} & $-$0.15 {\scriptsize [$-$0.17, $-$0.13]} & $-$0.01 {\scriptsize [$-$0.07, 0.04]} \\
UIConM~\cite{panetta2015human} & $-$0.09 {\scriptsize [$-$0.13, $-$0.04]} & 0.16 {\scriptsize [0.12, 0.20]} & 0.26 {\scriptsize [0.24, 0.28]} & 0.17 {\scriptsize [0.15, 0.18]} & 0.10 {\scriptsize [0.02, 0.19]} \\
UIQM~\cite{panetta2015human} & 0.18 {\scriptsize [0.13, 0.23]} & 0.29 {\scriptsize [0.24, 0.33]} & 0.34 {\scriptsize [0.33, 0.36]} & 0.18 {\scriptsize [0.16, 0.20]} & 0.15 {\scriptsize [0.06, 0.25]} \\
UISM~\cite{panetta2015human} & 0.25 {\scriptsize [0.20, 0.30]} & 0.23 {\scriptsize [0.19, 0.27]} & 0.24 {\scriptsize [0.22, 0.26]} & 0.17 {\scriptsize [0.16, 0.18]} & 0.19 {\scriptsize [0.10, 0.29]} \\
\multicolumn{6}{@{}l}{\emph{Learned underwater}} \\
NMFC~\cite{chen2023underwater} & 0.31 {\scriptsize [0.28, 0.35]} & 0.76 {\scriptsize [0.74, 0.79]} & 0.65 {\scriptsize [0.63, 0.66]} & 0.30 {\scriptsize [0.28, 0.31]} & 0.01 {\scriptsize [$-$0.10, 0.13]} \\
NR-UIQA~\cite{yi2024no} & 0.24 {\scriptsize [0.19, 0.29]} & 0.81 {\scriptsize [0.78, 0.83]} & 0.59 {\scriptsize [0.57, 0.61]} & 0.24 {\scriptsize [0.22, 0.26]} & 0.22 {\scriptsize [0.13, 0.31]} \\
NUIQ~\cite{jiang2022underwater} & 0.22 {\scriptsize [0.19, 0.26]} & 0.60 {\scriptsize [0.57, 0.63]} & 0.58 {\scriptsize [0.55, 0.60]} & 0.21 {\scriptsize [0.19, 0.23]} & 0.18 {\scriptsize [0.09, 0.27]} \\
QUIQ~\cite{karimi2025quiq} & 0.70 {\scriptsize [0.66, 0.74]} & 0.25 {\scriptsize [0.21, 0.29]} & 0.23 {\scriptsize [0.20, 0.25]} & 0.11 {\scriptsize [0.10, 0.13]} & 0.24 {\scriptsize [0.15, 0.33]} \\
Twice-Mixing~\cite{fu2022twice} & 0.33 {\scriptsize [0.29, 0.38]} & 0.38 {\scriptsize [0.34, 0.42]} & 0.38 {\scriptsize [0.35, 0.41]} & 0.19 {\scriptsize [0.17, 0.21]} & 0.30 {\scriptsize [0.21, 0.39]} \\
UIQAVSI~\cite{chen2025underwater} & 0.18 {\scriptsize [0.13, 0.22]} & 0.92 {\scriptsize [0.90, 0.95]} & 0.61 {\scriptsize [0.58, 0.63]} & 0.17 {\scriptsize [0.16, 0.19]} & 0.19 {\scriptsize [0.10, 0.29]} \\
URanker~\cite{guo2023underwater} & 0.41 {\scriptsize [0.37, 0.46]} & 0.20 {\scriptsize [0.16, 0.24]} & 0.21 {\scriptsize [0.19, 0.24]} & 0.27 {\scriptsize [0.25, 0.28]} & 0.06 {\scriptsize [$-$0.03, 0.15]} \\
\multicolumn{6}{@{}l}{\emph{Vision-language}} \\
Q-Align~\cite{wu2023q} & 0.23 {\scriptsize [0.17, 0.28]} & 0.53 {\scriptsize [0.49, 0.57]} & 0.45 {\scriptsize [0.42, 0.47]} & 0.24 {\scriptsize [0.22, 0.25]} & 0.35 {\scriptsize [0.26, 0.44]} \\
VisualQuality-R1~\cite{wu2026visualquality} & 0.19 {\scriptsize [0.13, 0.25]} & 0.52 {\scriptsize [0.47, 0.56]} & 0.50 {\scriptsize [0.47, 0.53]} & 0.26 {\scriptsize [0.24, 0.28]} & 0.44 {\scriptsize [0.33, 0.54]} \\
\multicolumn{6}{@{}l}{\emph{General-purpose}} \\
A-FINE~\cite{chen2025toward} & 0.23 {\scriptsize [0.17, 0.29]} & 0.29 {\scriptsize [0.22, 0.35]} & 0.36 {\scriptsize [0.33, 0.40]} & 0.10 {\scriptsize [0.08, 0.12]} & 0.37 {\scriptsize [0.29, 0.46]} \\
ARNIQA~\cite{agnolucci2024arniqa} & 0.18 {\scriptsize [0.13, 0.23]} & 0.26 {\scriptsize [0.21, 0.30]} & 0.22 {\scriptsize [0.19, 0.25]} & 0.19 {\scriptsize [0.18, 0.21]} & 0.25 {\scriptsize [0.16, 0.34]} \\
ARNIQA-CLIVE~\cite{agnolucci2024arniqa} & 0.27 {\scriptsize [0.22, 0.32]} & 0.11 {\scriptsize [0.07, 0.15]} & $-$0.01 {\scriptsize [$-$0.04, 0.02]} & 0.02 {\scriptsize [0.01, 0.04]} & $-$0.07 {\scriptsize [$-$0.16, 0.02]} \\
ARNIQA-CSIQ~\cite{agnolucci2024arniqa} & 0.10 {\scriptsize [0.06, 0.15]} & 0.28 {\scriptsize [0.24, 0.32]} & 0.26 {\scriptsize [0.23, 0.29]} & 0.27 {\scriptsize [0.25, 0.28]} & 0.26 {\scriptsize [0.19, 0.33]} \\
ARNIQA-FLIVE~\cite{agnolucci2024arniqa} & 0.16 {\scriptsize [0.12, 0.21]} & 0.22 {\scriptsize [0.17, 0.28]} & 0.15 {\scriptsize [0.12, 0.18]} & 0.05 {\scriptsize [0.04, 0.07]} & 0.17 {\scriptsize [0.07, 0.27]} \\
ARNIQA-KADID~\cite{agnolucci2024arniqa} & 0.13 {\scriptsize [0.09, 0.17]} & 0.35 {\scriptsize [0.31, 0.39]} & 0.21 {\scriptsize [0.18, 0.24]} & 0.19 {\scriptsize [0.17, 0.21]} & 0.29 {\scriptsize [0.20, 0.37]} \\
ARNIQA-LIVE~\cite{agnolucci2024arniqa} & 0.04 {\scriptsize [$-$0.01, 0.08]} & 0.41 {\scriptsize [0.37, 0.45]} & 0.32 {\scriptsize [0.29, 0.35]} & 0.31 {\scriptsize [0.29, 0.32]} & 0.25 {\scriptsize [0.18, 0.32]} \\
ARNIQA-SPAQ~\cite{agnolucci2024arniqa} & 0.14 {\scriptsize [0.09, 0.20]} & 0.24 {\scriptsize [0.19, 0.28]} & 0.21 {\scriptsize [0.18, 0.24]} & 0.14 {\scriptsize [0.13, 0.16]} & $-$0.07 {\scriptsize [$-$0.17, 0.03]} \\
ARNIQA-TID~\cite{agnolucci2024arniqa} & 0.00 {\scriptsize [$-$0.05, 0.05]} & 0.37 {\scriptsize [0.33, 0.41]} & 0.35 {\scriptsize [0.31, 0.38]} & 0.12 {\scriptsize [0.11, 0.14]} & 0.25 {\scriptsize [0.18, 0.32]} \\
BRISQUE~\cite{mittal2012no} & $-$0.04 {\scriptsize [$-$0.10, 0.02]} & 0.30 {\scriptsize [0.26, 0.35]} & 0.21 {\scriptsize [0.18, 0.24]} & 0.06 {\scriptsize [0.05, 0.08]} & 0.28 {\scriptsize [0.18, 0.38]} \\
BRISQUE (MATLAB)~\cite{mittal2012no} & 0.01 {\scriptsize [$-$0.04, 0.06]} & 0.17 {\scriptsize [0.12, 0.23]} & 0.10 {\scriptsize [0.06, 0.13]} & 0.03 {\scriptsize [0.02, 0.05]} & 0.22 {\scriptsize [0.13, 0.32]} \\
CLIP-IQA~\cite{wang2023exploring} & 0.00 {\scriptsize [$-$0.05, 0.05]} & 0.19 {\scriptsize [0.14, 0.24]} & 0.31 {\scriptsize [0.28, 0.34]} & $-$0.03 {\scriptsize [$-$0.05, $-$0.01]} & 0.11 {\scriptsize [0.02, 0.21]} \\
CLIP-IQA+~\cite{wang2023exploring} & 0.02 {\scriptsize [$-$0.03, 0.07]} & 0.33 {\scriptsize [0.28, 0.37]} & 0.36 {\scriptsize [0.33, 0.39]} & 0.06 {\scriptsize [0.05, 0.08]} & 0.29 {\scriptsize [0.20, 0.38]} \\
CLIP-IQA+ RN50~\cite{wang2023exploring} & $-$0.08 {\scriptsize [$-$0.13, $-$0.02]} & 0.23 {\scriptsize [0.17, 0.28]} & 0.27 {\scriptsize [0.24, 0.29]} & 0.03 {\scriptsize [0.02, 0.05]} & 0.24 {\scriptsize [0.15, 0.35]} \\
CLIP-IQA+ ViT-L~\cite{wang2023exploring} & 0.04 {\scriptsize [$-$0.01, 0.10]} & 0.16 {\scriptsize [0.11, 0.21]} & 0.20 {\scriptsize [0.16, 0.23]} & 0.08 {\scriptsize [0.07, 0.10]} & 0.15 {\scriptsize [0.06, 0.26]} \\
CNNIQA~\cite{kang2014convolutional} & 0.38 {\scriptsize [0.34, 0.42]} & 0.22 {\scriptsize [0.17, 0.27]} & 0.10 {\scriptsize [0.07, 0.13]} & 0.07 {\scriptsize [0.06, 0.09]} & $-$0.07 {\scriptsize [$-$0.15, 0.01]} \\
DBCNN~\cite{zhang2018blind} & 0.19 {\scriptsize [0.14, 0.24]} & 0.23 {\scriptsize [0.18, 0.27]} & 0.00 {\scriptsize [$-$0.04, 0.03]} & 0.02 {\scriptsize [0.00, 0.04]} & 0.06 {\scriptsize [$-$0.05, 0.17]} \\
Entropy & 0.39 {\scriptsize [0.34, 0.43]} & 0.07 {\scriptsize [0.04, 0.11]} & 0.08 {\scriptsize [0.05, 0.11]} & 0.01 {\scriptsize [0.00, 0.03]} & 0.15 {\scriptsize [0.06, 0.24]} \\
HyperIQA~\cite{su2020blindly} & 0.04 {\scriptsize [$-$0.01, 0.09]} & 0.18 {\scriptsize [0.13, 0.24]} & 0.14 {\scriptsize [0.10, 0.18]} & 0.14 {\scriptsize [0.13, 0.16]} & 0.23 {\scriptsize [0.13, 0.32]} \\
IL-NIQE~\cite{zhang2015feature} & 0.28 {\scriptsize [0.24, 0.33]} & 0.29 {\scriptsize [0.25, 0.33]} & 0.28 {\scriptsize [0.26, 0.31]} & 0.18 {\scriptsize [0.16, 0.20]} & 0.31 {\scriptsize [0.22, 0.40]} \\
LAION-Aes~\cite{schuhmann2024christophschuhmann} & 0.22 {\scriptsize [0.17, 0.27]} & 0.22 {\scriptsize [0.17, 0.26]} & 0.12 {\scriptsize [0.09, 0.15]} & 0.04 {\scriptsize [0.02, 0.06]} & 0.40 {\scriptsize [0.31, 0.47]} \\
LIQE~\cite{zhang2023blind} & 0.23 {\scriptsize [0.18, 0.29]} & 0.39 {\scriptsize [0.33, 0.45]} & 0.39 {\scriptsize [0.36, 0.42]} & 0.22 {\scriptsize [0.20, 0.23]} & 0.42 {\scriptsize [0.32, 0.51]} \\
LIQE-mix~\cite{zhang2023blind} & 0.24 {\scriptsize [0.18, 0.29]} & 0.44 {\scriptsize [0.40, 0.48]} & 0.45 {\scriptsize [0.42, 0.47]} & n/a & n/a \\
MA-CLIP~\cite{liao2026beyond} & 0.02 {\scriptsize [$-$0.03, 0.07]} & 0.27 {\scriptsize [0.22, 0.31]} & 0.37 {\scriptsize [0.34, 0.39]} & 0.03 {\scriptsize [0.01, 0.05]} & 0.22 {\scriptsize [0.13, 0.32]} \\
MANIQA~\cite{yang2022maniqa} & $-$0.08 {\scriptsize [$-$0.13, $-$0.03]} & 0.16 {\scriptsize [0.12, 0.20]} & 0.04 {\scriptsize [0.01, 0.07]} & 0.00 {\scriptsize [$-$0.02, 0.02]} & 0.16 {\scriptsize [0.06, 0.25]} \\
MANIQA-KADID~\cite{yang2022maniqa} & 0.00 {\scriptsize [$-$0.04, 0.04]} & 0.26 {\scriptsize [0.22, 0.29]} & 0.34 {\scriptsize [0.31, 0.36]} & 0.17 {\scriptsize [0.16, 0.19]} & 0.29 {\scriptsize [0.19, 0.38]} \\
MANIQA-PIPAL~\cite{yang2022maniqa} & 0.05 {\scriptsize [0.00, 0.10]} & 0.39 {\scriptsize [0.35, 0.44]} & 0.29 {\scriptsize [0.26, 0.32]} & 0.18 {\scriptsize [0.17, 0.20]} & 0.27 {\scriptsize [0.19, 0.35]} \\
MetaIQA~\cite{zhu2020metaiqa} & 0.28 {\scriptsize [0.23, 0.33]} & 0.20 {\scriptsize [0.15, 0.24]} & 0.21 {\scriptsize [0.18, 0.25]} & 0.11 {\scriptsize [0.09, 0.13]} & 0.09 {\scriptsize [0.00, 0.17]} \\
MUSIQ~\cite{ke2021musiq} & 0.16 {\scriptsize [0.12, 0.21]} & 0.38 {\scriptsize [0.33, 0.43]} & 0.33 {\scriptsize [0.30, 0.36]} & 0.11 {\scriptsize [0.09, 0.13]} & 0.09 {\scriptsize [0.00, 0.19]} \\
MUSIQ-AVA~\cite{ke2021musiq} & 0.17 {\scriptsize [0.11, 0.22]} & 0.29 {\scriptsize [0.24, 0.34]} & 0.14 {\scriptsize [0.11, 0.18]} & 0.17 {\scriptsize [0.15, 0.19]} & 0.35 {\scriptsize [0.27, 0.44]} \\
MUSIQ-PaQ2PiQ~\cite{ke2021musiq} & 0.22 {\scriptsize [0.16, 0.27]} & 0.28 {\scriptsize [0.23, 0.34]} & 0.21 {\scriptsize [0.18, 0.25]} & 0.05 {\scriptsize [0.03, 0.07]} & 0.18 {\scriptsize [0.09, 0.28]} \\
MUSIQ-SPAQ~\cite{ke2021musiq} & 0.14 {\scriptsize [0.09, 0.19]} & 0.25 {\scriptsize [0.20, 0.29]} & 0.19 {\scriptsize [0.16, 0.23]} & 0.03 {\scriptsize [0.01, 0.05]} & 0.02 {\scriptsize [$-$0.07, 0.12]} \\
NIMA~\cite{talebi2018nima} & 0.06 {\scriptsize [0.00, 0.11]} & 0.16 {\scriptsize [0.11, 0.22]} & 0.04 {\scriptsize [0.01, 0.07]} & 0.02 {\scriptsize [0.00, 0.04]} & 0.22 {\scriptsize [0.13, 0.31]} \\
NIMA-KONIQ~\cite{talebi2018nima} & 0.29 {\scriptsize [0.24, 0.34]} & 0.41 {\scriptsize [0.36, 0.45]} & 0.26 {\scriptsize [0.23, 0.29]} & 0.15 {\scriptsize [0.13, 0.17]} & 0.17 {\scriptsize [0.08, 0.26]} \\
NIMA-SPAQ~\cite{talebi2018nima} & 0.36 {\scriptsize [0.31, 0.41]} & 0.31 {\scriptsize [0.27, 0.35]} & 0.19 {\scriptsize [0.16, 0.22]} & 0.01 {\scriptsize [0.00, 0.03]} & 0.02 {\scriptsize [$-$0.06, 0.11]} \\
NIMA-VGG16-AVA~\cite{talebi2018nima} & 0.04 {\scriptsize [$-$0.01, 0.09]} & 0.19 {\scriptsize [0.14, 0.24]} & 0.14 {\scriptsize [0.10, 0.17]} & 0.01 {\scriptsize [$-$0.01, 0.02]} & 0.16 {\scriptsize [0.06, 0.26]} \\
NIQE~\cite{mittal2012making} & 0.05 {\scriptsize [0.00, 0.11]} & 0.11 {\scriptsize [0.06, 0.17]} & $-$0.12 {\scriptsize [$-$0.15, $-$0.09]} & 0.01 {\scriptsize [0.00, 0.03]} & 0.34 {\scriptsize [0.26, 0.42]} \\
NIQE (MATLAB)~\cite{mittal2012making} & $-$0.14 {\scriptsize [$-$0.18, $-$0.10]} & 0.00 {\scriptsize [$-$0.06, 0.05]} & $-$0.17 {\scriptsize [$-$0.21, $-$0.13]} & 0.07 {\scriptsize [0.05, 0.08]} & 0.28 {\scriptsize [0.17, 0.38]} \\
NRQM~\cite{ma2017learning} & 0.14 {\scriptsize [0.07, 0.21]} & 0.16 {\scriptsize [0.11, 0.21]} & 0.17 {\scriptsize [0.14, 0.20]} & $-$0.01 {\scriptsize [$-$0.02, 0.01]} & 0.08 {\scriptsize [$-$0.02, 0.19]} \\
PaQ-2-PiQ~\cite{ying2020patches} & 0.12 {\scriptsize [0.05, 0.18]} & 0.11 {\scriptsize [0.05, 0.16]} & 0.12 {\scriptsize [0.08, 0.16]} & $-$0.09 {\scriptsize [$-$0.11, $-$0.07]} & 0.12 {\scriptsize [0.04, 0.19]} \\
PI~\cite{blau20182018} & 0.21 {\scriptsize [0.15, 0.27]} & 0.18 {\scriptsize [0.13, 0.23]} & 0.07 {\scriptsize [0.04, 0.10]} & 0.02 {\scriptsize [0.00, 0.03]} & 0.34 {\scriptsize [0.26, 0.43]} \\
PIQE~\cite{venkatanath2015blind} & $-$0.10 {\scriptsize [$-$0.15, $-$0.05]} & 0.15 {\scriptsize [0.11, 0.20]} & 0.27 {\scriptsize [0.24, 0.30]} & 0.06 {\scriptsize [0.05, 0.08]} & 0.27 {\scriptsize [0.17, 0.36]} \\
QualiCLIP~\cite{agnolucci2024quality} & $-$0.01 {\scriptsize [$-$0.06, 0.03]} & 0.38 {\scriptsize [0.33, 0.43]} & 0.45 {\scriptsize [0.42, 0.48]} & 0.11 {\scriptsize [0.09, 0.12]} & 0.16 {\scriptsize [0.07, 0.27]} \\
QualiCLIP+~\cite{agnolucci2024quality} & 0.05 {\scriptsize [$-$0.01, 0.10]} & 0.31 {\scriptsize [0.27, 0.36]} & 0.30 {\scriptsize [0.27, 0.33]} & 0.14 {\scriptsize [0.12, 0.15]} & 0.13 {\scriptsize [0.03, 0.23]} \\
QualiCLIP+-CLIVE~\cite{agnolucci2024quality} & 0.00 {\scriptsize [$-$0.05, 0.05]} & 0.36 {\scriptsize [0.31, 0.41]} & 0.44 {\scriptsize [0.42, 0.47]} & 0.07 {\scriptsize [0.05, 0.08]} & 0.25 {\scriptsize [0.13, 0.36]} \\
QualiCLIP+-FLIVE~\cite{agnolucci2024quality} & 0.13 {\scriptsize [0.08, 0.18]} & 0.24 {\scriptsize [0.19, 0.29]} & 0.25 {\scriptsize [0.22, 0.29]} & $-$0.06 {\scriptsize [$-$0.08, $-$0.04]} & $-$0.04 {\scriptsize [$-$0.15, 0.07]} \\
QualiCLIP+-SPAQ~\cite{agnolucci2024quality} & $-$0.09 {\scriptsize [$-$0.14, $-$0.04]} & 0.30 {\scriptsize [0.26, 0.34]} & 0.39 {\scriptsize [0.36, 0.41]} & 0.05 {\scriptsize [0.03, 0.06]} & 0.32 {\scriptsize [0.22, 0.42]} \\
RALI~\cite{zhao2026reasoning} & 0.18 {\scriptsize [0.12, 0.23]} & 0.56 {\scriptsize [0.52, 0.59]} & 0.46 {\scriptsize [0.44, 0.49]} & 0.24 {\scriptsize [0.22, 0.26]} & 0.33 {\scriptsize [0.25, 0.42]} \\
TOPIQ-IAA~\cite{chen2024topiq} & 0.16 {\scriptsize [0.10, 0.22]} & 0.31 {\scriptsize [0.26, 0.36]} & 0.13 {\scriptsize [0.10, 0.16]} & 0.12 {\scriptsize [0.11, 0.14]} & 0.31 {\scriptsize [0.20, 0.40]} \\
TOPIQ-IAA R50~\cite{chen2024topiq} & 0.02 {\scriptsize [$-$0.04, 0.07]} & 0.22 {\scriptsize [0.18, 0.27]} & 0.10 {\scriptsize [0.07, 0.13]} & 0.13 {\scriptsize [0.11, 0.15]} & 0.26 {\scriptsize [0.18, 0.35]} \\
TOPIQ-NR~\cite{chen2024topiq} & 0.06 {\scriptsize [0.01, 0.11]} & 0.28 {\scriptsize [0.23, 0.32]} & 0.19 {\scriptsize [0.16, 0.22]} & 0.13 {\scriptsize [0.11, 0.15]} & 0.28 {\scriptsize [0.18, 0.38]} \\
TOPIQ-NR-FLIVE~\cite{chen2024topiq} & 0.19 {\scriptsize [0.13, 0.25]} & 0.22 {\scriptsize [0.17, 0.27]} & 0.21 {\scriptsize [0.18, 0.24]} & 0.02 {\scriptsize [0.00, 0.04]} & 0.19 {\scriptsize [0.10, 0.27]} \\
TOPIQ-NR-SPAQ~\cite{chen2024topiq} & 0.07 {\scriptsize [0.01, 0.13]} & 0.22 {\scriptsize [0.17, 0.28]} & 0.17 {\scriptsize [0.13, 0.20]} & 0.01 {\scriptsize [$-$0.01, 0.03]} & 0.28 {\scriptsize [0.22, 0.36]} \\
TReS~\cite{golestaneh2022no} & 0.01 {\scriptsize [$-$0.04, 0.06]} & 0.11 {\scriptsize [0.07, 0.15]} & 0.13 {\scriptsize [0.10, 0.16]} & 0.16 {\scriptsize [0.14, 0.18]} & 0.16 {\scriptsize [0.06, 0.26]} \\
TReS-FLIVE~\cite{golestaneh2022no} & 0.10 {\scriptsize [0.04, 0.17]} & 0.21 {\scriptsize [0.16, 0.28]} & 0.26 {\scriptsize [0.22, 0.29]} & 0.10 {\scriptsize [0.08, 0.11]} & 0.16 {\scriptsize [0.06, 0.25]} \\
UNIQUE~\cite{zhang2021uncertainty} & 0.15 {\scriptsize [0.09, 0.20]} & 0.38 {\scriptsize [0.34, 0.42]} & 0.37 {\scriptsize [0.33, 0.40]} & 0.25 {\scriptsize [0.23, 0.27]} & 0.29 {\scriptsize [0.20, 0.38]} \\
WaDIQaM-NR~\cite{bosse2017deep} & 0.21 {\scriptsize [0.16, 0.27]} & 0.10 {\scriptsize [0.06, 0.15]} & 0.01 {\scriptsize [$-$0.02, 0.04]} & 0.04 {\scriptsize [0.02, 0.05]} & 0.12 {\scriptsize [0.04, 0.21]} \\
\bottomrule
\end{longtable}

\begin{longtable}{@{}lcccccc@{}}
\caption{Share of scenes (\%) in which the simple edit of the raw photo scores above the people's best enhancement, all metrics, maximum over the three MOS sets and, in brackets, the set where it is reached. Values are rounded, so a printed 50 can be just above half. NUIQ and LIQE-mix were not run on the edits.}\label{tab:edits_all}\\
\toprule
Metric & S+S & S+S+C & S+S+N & CLAHE & Red boost & Noise \\
\midrule
\endfirsthead
\toprule
Metric & S+S & S+S+C & S+S+N & CLAHE & Red boost & Noise \\
\midrule
\endhead
A-FINE~\cite{chen2025toward} & 62 {\scriptsize (SOTA)} & 45 {\scriptsize (UID)} & 36 {\scriptsize (SAUD)} & 28 {\scriptsize (SOTA)} & 56 {\scriptsize (SOTA)} & 12 {\scriptsize (SAUD)} \\
ARNIQA~\cite{agnolucci2024arniqa} & 30 {\scriptsize (UID)} & 42 {\scriptsize (UID)} & 10 {\scriptsize (UID)} & 43 {\scriptsize (UID)} & 45 {\scriptsize (UID)} & 5 {\scriptsize (UID)} \\
ARNIQA-CLIVE~\cite{agnolucci2024arniqa} & 60 {\scriptsize (SOTA)} & 98 {\scriptsize (SOTA)} & 46 {\scriptsize (SOTA)} & 84 {\scriptsize (SOTA)} & 32 {\scriptsize (SOTA)} & 12 {\scriptsize (SOTA)} \\
ARNIQA-CSIQ~\cite{agnolucci2024arniqa} & 17 {\scriptsize (UID)} & 30 {\scriptsize (UID)} & 3 {\scriptsize (UID)} & 57 {\scriptsize (UID)} & 53 {\scriptsize (UID)} & 7 {\scriptsize (UID)} \\
ARNIQA-FLIVE~\cite{agnolucci2024arniqa} & 68 {\scriptsize (SOTA)} & 69 {\scriptsize (SAUD)} & 29 {\scriptsize (SAUD)} & 43 {\scriptsize (UID)} & 47 {\scriptsize (SOTA)} & 11 {\scriptsize (SOTA)} \\
ARNIQA-KADID~\cite{agnolucci2024arniqa} & 18 {\scriptsize (UID)} & 37 {\scriptsize (UID)} & 8 {\scriptsize (UID)} & 62 {\scriptsize (UID)} & 43 {\scriptsize (UID)} & 12 {\scriptsize (UID)} \\
ARNIQA-LIVE~\cite{agnolucci2024arniqa} & 25 {\scriptsize (UID)} & 25 {\scriptsize (UID)} & 4 {\scriptsize (SOTA)} & 42 {\scriptsize (UID)} & 68 {\scriptsize (UID)} & 5 {\scriptsize (UID)} \\
ARNIQA-SPAQ~\cite{agnolucci2024arniqa} & 42 {\scriptsize (SOTA)} & 61 {\scriptsize (SAUD)} & 45 {\scriptsize (UID)} & 63 {\scriptsize (SAUD)} & 47 {\scriptsize (UID)} & 25 {\scriptsize (SAUD)} \\
ARNIQA-TID~\cite{agnolucci2024arniqa} & 63 {\scriptsize (UID)} & 30 {\scriptsize (UID)} & 65 {\scriptsize (UID)} & 60 {\scriptsize (UID)} & 73 {\scriptsize (UID)} & 63 {\scriptsize (UID)} \\
BRISQUE~\cite{mittal2012no} & 60 {\scriptsize (UID)} & 27 {\scriptsize (SOTA)} & 10 {\scriptsize (SOTA)} & 31 {\scriptsize (SOTA)} & 78 {\scriptsize (UID)} & 9 {\scriptsize (SOTA)} \\
BRISQUE (MATLAB)~\cite{mittal2012no} & 68 {\scriptsize (SOTA)} & 38 {\scriptsize (SOTA)} & 22 {\scriptsize (SOTA)} & 54 {\scriptsize (SOTA)} & 57 {\scriptsize (UID)} & 20 {\scriptsize (SOTA)} \\
CCF~\cite{wang2018imaging} & 80 {\scriptsize (SOTA)} & 100 {\scriptsize (SOTA)} & 80 {\scriptsize (SOTA)} & 100 {\scriptsize (SOTA)} & 46 {\scriptsize (SOTA)} & 59 {\scriptsize (SOTA)} \\
CLIP-IQA~\cite{wang2023exploring} & 40 {\scriptsize (UID)} & 38 {\scriptsize (UID)} & 68 {\scriptsize (SOTA)} & 53 {\scriptsize (UID)} & 74 {\scriptsize (SOTA)} & 47 {\scriptsize (SOTA)} \\
CLIP-IQA+~\cite{wang2023exploring} & 42 {\scriptsize (UID)} & 22 {\scriptsize (UID)} & 20 {\scriptsize (SOTA)} & 63 {\scriptsize (UID)} & 68 {\scriptsize (SOTA)} & 10 {\scriptsize (SOTA)} \\
CLIP-IQA+ RN50~\cite{wang2023exploring} & 68 {\scriptsize (UID)} & 37 {\scriptsize (UID)} & 38 {\scriptsize (UID)} & 67 {\scriptsize (UID)} & 73 {\scriptsize (UID)} & 28 {\scriptsize (UID)} \\
CLIP-IQA+ ViT-L~\cite{wang2023exploring} & 40 {\scriptsize (SAUD)} & 43 {\scriptsize (UID)} & 23 {\scriptsize (UID)} & 51 {\scriptsize (SAUD)} & 46 {\scriptsize (SOTA)} & 12 {\scriptsize (UID)} \\
CNNIQA~\cite{kang2014convolutional} & 64 {\scriptsize (SAUD)} & 74 {\scriptsize (SAUD)} & 67 {\scriptsize (SAUD)} & 71 {\scriptsize (SAUD)} & 26 {\scriptsize (SAUD)} & 24 {\scriptsize (SAUD)} \\
DBCNN~\cite{zhang2018blind} & 60 {\scriptsize (SOTA)} & 82 {\scriptsize (UID)} & 42 {\scriptsize (SAUD)} & 63 {\scriptsize (UID)} & 31 {\scriptsize (SOTA)} & 28 {\scriptsize (SOTA)} \\
Entropy & 20 {\scriptsize (SOTA)} & 74 {\scriptsize (SOTA)} & 30 {\scriptsize (SOTA)} & 74 {\scriptsize (SOTA)} & 14 {\scriptsize (SOTA)} & 18 {\scriptsize (SOTA)} \\
FDUM~\cite{yang2021reference} & 78 {\scriptsize (SOTA)} & 99 {\scriptsize (SOTA)} & 97 {\scriptsize (SOTA)} & 88 {\scriptsize (SOTA)} & 41 {\scriptsize (SOTA)} & 86 {\scriptsize (SOTA)} \\
HyperIQA~\cite{su2020blindly} & 70 {\scriptsize (UID)} & 78 {\scriptsize (UID)} & 17 {\scriptsize (SAUD)} & 67 {\scriptsize (UID)} & 52 {\scriptsize (SOTA)} & 2 {\scriptsize (SOTA)} \\
IL-NIQE~\cite{zhang2015feature} & 15 {\scriptsize (SOTA)} & 16 {\scriptsize (SOTA)} & 2 {\scriptsize (SAUD)} & 37 {\scriptsize (SOTA)} & 15 {\scriptsize (SAUD)} & 4 {\scriptsize (SAUD)} \\
LAION-Aes~\cite{schuhmann2024christophschuhmann} & 68 {\scriptsize (SOTA)} & 72 {\scriptsize (SAUD)} & 54 {\scriptsize (SOTA)} & 69 {\scriptsize (SAUD)} & 51 {\scriptsize (SOTA)} & 22 {\scriptsize (SOTA)} \\
LIQE~\cite{zhang2023blind} & 26 {\scriptsize (SOTA)} & 33 {\scriptsize (UID)} & 6 {\scriptsize (SOTA)} & 44 {\scriptsize (SOTA)} & 31 {\scriptsize (SOTA)} & 4 {\scriptsize (SOTA)} \\
MA-CLIP~\cite{liao2026beyond} & 40 {\scriptsize (UID)} & 38 {\scriptsize (UID)} & 58 {\scriptsize (SOTA)} & 63 {\scriptsize (UID)} & 72 {\scriptsize (SOTA)} & 28 {\scriptsize (SOTA)} \\
MANIQA~\cite{yang2022maniqa} & 85 {\scriptsize (UID)} & 80 {\scriptsize (UID)} & 10 {\scriptsize (UID)} & 77 {\scriptsize (UID)} & 75 {\scriptsize (UID)} & 6 {\scriptsize (SAUD)} \\
MANIQA-KADID~\cite{yang2022maniqa} & 47 {\scriptsize (UID)} & 7 {\scriptsize (UID)} & 45 {\scriptsize (UID)} & 27 {\scriptsize (UID)} & 75 {\scriptsize (UID)} & 65 {\scriptsize (UID)} \\
MANIQA-PIPAL~\cite{yang2022maniqa} & 87 {\scriptsize (UID)} & 78 {\scriptsize (UID)} & 20 {\scriptsize (UID)} & 65 {\scriptsize (UID)} & 65 {\scriptsize (UID)} & 1 {\scriptsize (SAUD)} \\
MetaIQA~\cite{zhu2020metaiqa} & 60 {\scriptsize (SOTA)} & 71 {\scriptsize (SOTA)} & 58 {\scriptsize (SOTA)} & 62 {\scriptsize (SOTA)} & 32 {\scriptsize (SOTA)} & 20 {\scriptsize (SOTA)} \\
MUSIQ~\cite{ke2021musiq} & 69 {\scriptsize (SOTA)} & 73 {\scriptsize (UID)} & 20 {\scriptsize (SOTA)} & 70 {\scriptsize (UID)} & 42 {\scriptsize (SOTA)} & 5 {\scriptsize (UID)} \\
MUSIQ-AVA~\cite{ke2021musiq} & 43 {\scriptsize (UID)} & 35 {\scriptsize (UID)} & 25 {\scriptsize (UID)} & 59 {\scriptsize (SAUD)} & 35 {\scriptsize (UID)} & 16 {\scriptsize (SOTA)} \\
MUSIQ-PaQ2PiQ~\cite{ke2021musiq} & 79 {\scriptsize (SOTA)} & 80 {\scriptsize (SOTA)} & 62 {\scriptsize (SOTA)} & 46 {\scriptsize (SOTA)} & 41 {\scriptsize (SOTA)} & 12 {\scriptsize (SOTA)} \\
MUSIQ-SPAQ~\cite{ke2021musiq} & 50 {\scriptsize (SOTA)} & 67 {\scriptsize (UID)} & 37 {\scriptsize (UID)} & 55 {\scriptsize (UID)} & 49 {\scriptsize (SOTA)} & 12 {\scriptsize (SOTA)} \\
NIMA~\cite{talebi2018nima} & 51 {\scriptsize (SAUD)} & 56 {\scriptsize (SAUD)} & 40 {\scriptsize (UID)} & 75 {\scriptsize (UID)} & 47 {\scriptsize (UID)} & 23 {\scriptsize (UID)} \\
NIMA-KONIQ~\cite{talebi2018nima} & 72 {\scriptsize (SOTA)} & 75 {\scriptsize (SAUD)} & 22 {\scriptsize (SOTA)} & 38 {\scriptsize (SAUD)} & 32 {\scriptsize (SOTA)} & 2 {\scriptsize (UID)} \\
NIMA-SPAQ~\cite{talebi2018nima} & 86 {\scriptsize (SOTA)} & 96 {\scriptsize (SAUD)} & 84 {\scriptsize (SOTA)} & 44 {\scriptsize (SOTA)} & 32 {\scriptsize (SOTA)} & 13 {\scriptsize (SOTA)} \\
NIMA-VGG16-AVA~\cite{talebi2018nima} & 67 {\scriptsize (UID)} & 75 {\scriptsize (UID)} & 52 {\scriptsize (UID)} & 87 {\scriptsize (UID)} & 63 {\scriptsize (UID)} & 30 {\scriptsize (UID)} \\
NIQE~\cite{mittal2012making} & 45 {\scriptsize (SAUD)} & 50 {\scriptsize (SOTA)} & 2 {\scriptsize (SOTA)} & 76 {\scriptsize (SOTA)} & 42 {\scriptsize (UID)} & 0 {\scriptsize (SAUD)} \\
NIQE (MATLAB)~\cite{mittal2012making} & 48 {\scriptsize (UID)} & 26 {\scriptsize (SOTA)} & 0 {\scriptsize (SOTA)} & 55 {\scriptsize (SOTA)} & 77 {\scriptsize (UID)} & 0 {\scriptsize (SAUD)} \\
NMFC~\cite{chen2023underwater} & 8 {\scriptsize (UID)} & 8 {\scriptsize (UID)} & 22 {\scriptsize (SOTA)} & 19 {\scriptsize (SOTA)} & 16 {\scriptsize (SOTA)} & 16 {\scriptsize (SOTA)} \\
NR-UIQA~\cite{yi2024no} & 38 {\scriptsize (SOTA)} & 38 {\scriptsize (SOTA)} & 19 {\scriptsize (SOTA)} & 48 {\scriptsize (SOTA)} & 22 {\scriptsize (SOTA)} & 23 {\scriptsize (SOTA)} \\
NRQM~\cite{ma2017learning} & 57 {\scriptsize (SAUD)} & 41 {\scriptsize (SOTA)} & 51 {\scriptsize (SAUD)} & 32 {\scriptsize (UID)} & 54 {\scriptsize (SAUD)} & 52 {\scriptsize (SOTA)} \\
PaQ-2-PiQ~\cite{ying2020patches} & 97 {\scriptsize (SAUD)} & 94 {\scriptsize (SOTA)} & 88 {\scriptsize (SAUD)} & 73 {\scriptsize (SOTA)} & 69 {\scriptsize (SOTA)} & 25 {\scriptsize (UID)} \\
PI~\cite{blau20182018} & 55 {\scriptsize (SAUD)} & 48 {\scriptsize (SOTA)} & 8 {\scriptsize (SOTA)} & 57 {\scriptsize (UID)} & 44 {\scriptsize (SAUD)} & 2 {\scriptsize (SOTA)} \\
PIQE~\cite{venkatanath2015blind} & 87 {\scriptsize (UID)} & 42 {\scriptsize (SOTA)} & 15 {\scriptsize (SOTA)} & 60 {\scriptsize (UID)} & 68 {\scriptsize (UID)} & 11 {\scriptsize (SOTA)} \\
Q-Align~\cite{wu2023q} & 27 {\scriptsize (SOTA)} & 8 {\scriptsize (UID)} & 6 {\scriptsize (SOTA)} & 42 {\scriptsize (UID)} & 40 {\scriptsize (SOTA)} & 4 {\scriptsize (SOTA)} \\
QualiCLIP~\cite{agnolucci2024quality} & 52 {\scriptsize (UID)} & 32 {\scriptsize (UID)} & 12 {\scriptsize (SOTA)} & 57 {\scriptsize (UID)} & 63 {\scriptsize (UID)} & 6 {\scriptsize (SOTA)} \\
QualiCLIP+~\cite{agnolucci2024quality} & 40 {\scriptsize (UID)} & 32 {\scriptsize (UID)} & 2 {\scriptsize (SOTA)} & 50 {\scriptsize (UID)} & 50 {\scriptsize (SOTA)} & 2 {\scriptsize (SOTA)} \\
QualiCLIP+-CLIVE~\cite{agnolucci2024quality} & 62 {\scriptsize (UID)} & 45 {\scriptsize (UID)} & 34 {\scriptsize (SOTA)} & 57 {\scriptsize (UID)} & 71 {\scriptsize (SOTA)} & 24 {\scriptsize (SOTA)} \\
QualiCLIP+-FLIVE~\cite{agnolucci2024quality} & 65 {\scriptsize (SOTA)} & 60 {\scriptsize (SOTA)} & 29 {\scriptsize (SOTA)} & 57 {\scriptsize (SOTA)} & 72 {\scriptsize (SOTA)} & 7 {\scriptsize (SAUD)} \\
QualiCLIP+-SPAQ~\cite{agnolucci2024quality} & 67 {\scriptsize (UID)} & 42 {\scriptsize (UID)} & 48 {\scriptsize (SOTA)} & 77 {\scriptsize (UID)} & 82 {\scriptsize (UID)} & 38 {\scriptsize (SOTA)} \\
QUIQ~\cite{karimi2025quiq} & 32 {\scriptsize (SOTA)} & 58 {\scriptsize (SOTA)} & 70 {\scriptsize (SOTA)} & 62 {\scriptsize (SOTA)} & 14 {\scriptsize (SOTA)} & 74 {\scriptsize (SOTA)} \\
RALI~\cite{zhao2026reasoning} & 53 {\scriptsize (UID)} & 42 {\scriptsize (UID)} & 7 {\scriptsize (SOTA)} & 58 {\scriptsize (UID)} & 43 {\scriptsize (UID)} & 1 {\scriptsize (SAUD)} \\
TOPIQ-IAA~\cite{chen2024topiq} & 32 {\scriptsize (SOTA)} & 30 {\scriptsize (SAUD)} & 23 {\scriptsize (UID)} & 55 {\scriptsize (UID)} & 28 {\scriptsize (SOTA)} & 23 {\scriptsize (UID)} \\
TOPIQ-IAA R50~\cite{chen2024topiq} & 50 {\scriptsize (SOTA)} & 57 {\scriptsize (UID)} & 35 {\scriptsize (SAUD)} & 87 {\scriptsize (UID)} & 57 {\scriptsize (UID)} & 13 {\scriptsize (UID)} \\
TOPIQ-NR~\cite{chen2024topiq} & 78 {\scriptsize (UID)} & 78 {\scriptsize (UID)} & 22 {\scriptsize (UID)} & 65 {\scriptsize (UID)} & 60 {\scriptsize (UID)} & 5 {\scriptsize (SOTA)} \\
TOPIQ-NR-FLIVE~\cite{chen2024topiq} & 90 {\scriptsize (SOTA)} & 92 {\scriptsize (SOTA)} & 74 {\scriptsize (SOTA)} & 60 {\scriptsize (SOTA)} & 56 {\scriptsize (SOTA)} & 14 {\scriptsize (SOTA)} \\
TOPIQ-NR-SPAQ~\cite{chen2024topiq} & 85 {\scriptsize (SOTA)} & 94 {\scriptsize (SOTA)} & 66 {\scriptsize (SOTA)} & 85 {\scriptsize (UID)} & 64 {\scriptsize (SOTA)} & 13 {\scriptsize (UID)} \\
TReS~\cite{golestaneh2022no} & 68 {\scriptsize (UID)} & 75 {\scriptsize (UID)} & 7 {\scriptsize (UID)} & 73 {\scriptsize (UID)} & 60 {\scriptsize (UID)} & 0 {\scriptsize (SAUD)} \\
TReS-FLIVE~\cite{golestaneh2022no} & 66 {\scriptsize (SOTA)} & 53 {\scriptsize (UID)} & 30 {\scriptsize (UID)} & 40 {\scriptsize (UID)} & 62 {\scriptsize (SOTA)} & 18 {\scriptsize (UID)} \\
Twice-Mixing~\cite{fu2022twice} & 20 {\scriptsize (SOTA)} & 48 {\scriptsize (SOTA)} & 29 {\scriptsize (SOTA)} & 89 {\scriptsize (SOTA)} & 22 {\scriptsize (SOTA)} & 34 {\scriptsize (SOTA)} \\
UCIQE~\cite{yang2015underwater} & 81 {\scriptsize (SOTA)} & 95 {\scriptsize (SOTA)} & 88 {\scriptsize (SOTA)} & 76 {\scriptsize (SOTA)} & 56 {\scriptsize (SOTA)} & 72 {\scriptsize (SOTA)} \\
UICM~\cite{panetta2015human} & 68 {\scriptsize (SOTA)} & 72 {\scriptsize (SOTA)} & 76 {\scriptsize (SOTA)} & 42 {\scriptsize (SOTA)} & 64 {\scriptsize (SOTA)} & 68 {\scriptsize (SOTA)} \\
UIConM~\cite{panetta2015human} & 18 {\scriptsize (SAUD)} & 2 {\scriptsize (SOTA)} & 10 {\scriptsize (SAUD)} & 27 {\scriptsize (UID)} & 63 {\scriptsize (UID)} & 37 {\scriptsize (SAUD)} \\
UIQAVSI~\cite{chen2025underwater} & 22 {\scriptsize (SOTA)} & 36 {\scriptsize (SOTA)} & 29 {\scriptsize (SOTA)} & 56 {\scriptsize (SOTA)} & 28 {\scriptsize (UID)} & 51 {\scriptsize (SOTA)} \\
UIQM~\cite{panetta2015human} & 20 {\scriptsize (SAUD)} & 21 {\scriptsize (SOTA)} & 19 {\scriptsize (SAUD)} & 35 {\scriptsize (SOTA)} & 28 {\scriptsize (UID)} & 41 {\scriptsize (SAUD)} \\
UISM~\cite{panetta2015human} & 28 {\scriptsize (SOTA)} & 64 {\scriptsize (SOTA)} & 49 {\scriptsize (SOTA)} & 66 {\scriptsize (SOTA)} & 22 {\scriptsize (UID)} & 68 {\scriptsize (SOTA)} \\
UNIQUE~\cite{zhang2021uncertainty} & 38 {\scriptsize (UID)} & 48 {\scriptsize (UID)} & 20 {\scriptsize (UID)} & 60 {\scriptsize (UID)} & 40 {\scriptsize (UID)} & 15 {\scriptsize (UID)} \\
URanker~\cite{guo2023underwater} & 4 {\scriptsize (SOTA)} & 64 {\scriptsize (SOTA)} & 8 {\scriptsize (SOTA)} & 72 {\scriptsize (SOTA)} & 3 {\scriptsize (SOTA)} & 2 {\scriptsize (SOTA)} \\
VisualQuality-R1~\cite{wu2026visualquality} & 26 {\scriptsize (SOTA)} & 8 {\scriptsize (UID)} & 6 {\scriptsize (SOTA)} & 26 {\scriptsize (SOTA)} & 36 {\scriptsize (SOTA)} & 5 {\scriptsize (SOTA)} \\
WaDIQaM-NR~\cite{bosse2017deep} & 72 {\scriptsize (SAUD)} & 85 {\scriptsize (SAUD)} & 73 {\scriptsize (SAUD)} & 69 {\scriptsize (SAUD)} & 53 {\scriptsize (SAUD)} & 38 {\scriptsize (SAUD)} \\
\bottomrule
\end{longtable}

\begin{longtable}{@{}lcccc@{}}
\caption{Drift test for all 73 scores: 72 metrics (LIQE-mix was not run on the chains) and, as a diagnostic, the patched MATLAB UCIQE code. Each cell: mean change of the min-max normalized score from step 1 to the last step (negative = the score falls as the photo degrades), per-image pass rate in brackets, \cmark\ when the score detects the drift (below step 1 at every checked step).}\label{tab:drift_all}\\
\toprule
Score & Pro replica & NB2 replica & Pro enhance & NB2 enhance \\
\midrule
\endfirsthead
\toprule
Score & Pro replica & NB2 replica & Pro enhance & NB2 enhance \\
\midrule
\endhead
A-FINE~\cite{chen2025toward} & 0.04 {\scriptsize (20)} & \cmark\ $-$0.07 {\scriptsize (40)} & \cmark\ $-$0.33 {\scriptsize (50)} & \cmark\ $-$0.29 {\scriptsize (75)} \\
ARNIQA~\cite{agnolucci2024arniqa} & \cmark\ $-$0.40 {\scriptsize (70)} & \cmark\ $-$0.27 {\scriptsize (40)} & \cmark\ $-$0.52 {\scriptsize (70)} & $-$0.38 {\scriptsize (40)} \\
ARNIQA-CLIVE~\cite{agnolucci2024arniqa} & 0.34 {\scriptsize (0)} & 0.31 {\scriptsize (5)} & 0.13 {\scriptsize (25)} & \cmark\ $-$0.17 {\scriptsize (60)} \\
ARNIQA-CSIQ~\cite{agnolucci2024arniqa} & \cmark\ $-$0.44 {\scriptsize (55)} & $-$0.37 {\scriptsize (35)} & \cmark\ $-$0.59 {\scriptsize (95)} & \cmark\ $-$0.53 {\scriptsize (90)} \\
ARNIQA-FLIVE~\cite{agnolucci2024arniqa} & $-$0.08 {\scriptsize (35)} & \cmark\ $-$0.04 {\scriptsize (25)} & 0.16 {\scriptsize (5)} & \cmark\ $-$0.11 {\scriptsize (50)} \\
ARNIQA-KADID~\cite{agnolucci2024arniqa} & $-$0.38 {\scriptsize (30)} & $-$0.32 {\scriptsize (15)} & \cmark\ $-$0.51 {\scriptsize (85)} & \cmark\ $-$0.52 {\scriptsize (90)} \\
ARNIQA-LIVE~\cite{agnolucci2024arniqa} & \cmark\ $-$0.46 {\scriptsize (65)} & $-$0.40 {\scriptsize (40)} & \cmark\ $-$0.58 {\scriptsize (95)} & \cmark\ $-$0.58 {\scriptsize (95)} \\
ARNIQA-SPAQ~\cite{agnolucci2024arniqa} & 0.25 {\scriptsize (0)} & 0.09 {\scriptsize (15)} & \cmark\ $-$0.16 {\scriptsize (80)} & \cmark\ $-$0.27 {\scriptsize (70)} \\
ARNIQA-TID~\cite{agnolucci2024arniqa} & \cmark\ $-$0.46 {\scriptsize (95)} & \cmark\ $-$0.42 {\scriptsize (65)} & \cmark\ $-$0.57 {\scriptsize (100)} & \cmark\ $-$0.57 {\scriptsize (95)} \\
BRISQUE~\cite{mittal2012no} & $-$0.20 {\scriptsize (10)} & 0.13 {\scriptsize (10)} & 0.36 {\scriptsize (5)} & 0.27 {\scriptsize (10)} \\
BRISQUE (MATLAB)~\cite{mittal2012no} & $-$0.12 {\scriptsize (20)} & 0.23 {\scriptsize (10)} & 0.36 {\scriptsize (0)} & 0.22 {\scriptsize (10)} \\
CCF~\cite{wang2018imaging} & 0.45 {\scriptsize (5)} & 0.44 {\scriptsize (0)} & 0.11 {\scriptsize (10)} & 0.14 {\scriptsize (5)} \\
CLIP-IQA~\cite{wang2023exploring} & 0.26 {\scriptsize (0)} & 0.27 {\scriptsize (5)} & 0.26 {\scriptsize (5)} & 0.03 {\scriptsize (20)} \\
CLIP-IQA+~\cite{wang2023exploring} & 0.05 {\scriptsize (5)} & 0.09 {\scriptsize (15)} & 0.19 {\scriptsize (5)} & $-$0.01 {\scriptsize (20)} \\
CLIP-IQA+ RN50~\cite{wang2023exploring} & 0.44 {\scriptsize (0)} & 0.28 {\scriptsize (10)} & 0.07 {\scriptsize (20)} & $-$0.12 {\scriptsize (50)} \\
CLIP-IQA+ ViT-L~\cite{wang2023exploring} & 0.19 {\scriptsize (10)} & 0.34 {\scriptsize (15)} & 0.22 {\scriptsize (10)} & 0.05 {\scriptsize (25)} \\
CNNIQA~\cite{kang2014convolutional} & 0.56 {\scriptsize (0)} & 0.57 {\scriptsize (0)} & 0.43 {\scriptsize (0)} & 0.30 {\scriptsize (0)} \\
DBCNN~\cite{zhang2018blind} & 0.48 {\scriptsize (0)} & 0.35 {\scriptsize (5)} & 0.39 {\scriptsize (5)} & 0.28 {\scriptsize (15)} \\
Entropy & \cmark\ $-$0.04 {\scriptsize (40)} & 0.03 {\scriptsize (30)} & 0.07 {\scriptsize (0)} & 0.07 {\scriptsize (15)} \\
FDUM~\cite{yang2021reference} & 0.33 {\scriptsize (0)} & 0.39 {\scriptsize (5)} & 0.40 {\scriptsize (0)} & 0.26 {\scriptsize (0)} \\
HyperIQA~\cite{su2020blindly} & 0.45 {\scriptsize (0)} & 0.38 {\scriptsize (5)} & 0.25 {\scriptsize (10)} & 0.12 {\scriptsize (10)} \\
IL-NIQE~\cite{zhang2015feature} & \cmark\ $-$0.20 {\scriptsize (45)} & $-$0.22 {\scriptsize (40)} & \cmark\ $-$0.29 {\scriptsize (50)} & $-$0.15 {\scriptsize (45)} \\
LAION-Aes~\cite{schuhmann2024christophschuhmann} & \cmark\ $-$0.52 {\scriptsize (75)} & $-$0.30 {\scriptsize (35)} & $-$0.29 {\scriptsize (40)} & $-$0.17 {\scriptsize (40)} \\
LIQE~\cite{zhang2023blind} & \cmark\ $-$0.10 {\scriptsize (35)} & 0.01 {\scriptsize (15)} & $-$0.05 {\scriptsize (5)} & $-$0.17 {\scriptsize (30)} \\
MA-CLIP~\cite{liao2026beyond} & 0.07 {\scriptsize (0)} & 0.09 {\scriptsize (10)} & 0.17 {\scriptsize (15)} & \cmark\ $-$0.03 {\scriptsize (35)} \\
MANIQA~\cite{yang2022maniqa} & 0.35 {\scriptsize (10)} & 0.25 {\scriptsize (10)} & 0.15 {\scriptsize (25)} & 0.04 {\scriptsize (25)} \\
MANIQA-KADID~\cite{yang2022maniqa} & $-$0.30 {\scriptsize (15)} & $-$0.33 {\scriptsize (10)} & $-$0.37 {\scriptsize (15)} & $-$0.21 {\scriptsize (15)} \\
MANIQA-PIPAL~\cite{yang2022maniqa} & \cmark\ $-$0.23 {\scriptsize (90)} & \cmark\ $-$0.19 {\scriptsize (50)} & \cmark\ $-$0.43 {\scriptsize (85)} & \cmark\ $-$0.45 {\scriptsize (90)} \\
MetaIQA~\cite{zhu2020metaiqa} & 0.03 {\scriptsize (35)} & 0.08 {\scriptsize (30)} & 0.07 {\scriptsize (15)} & $-$0.02 {\scriptsize (25)} \\
MUSIQ~\cite{ke2021musiq} & 0.23 {\scriptsize (0)} & 0.21 {\scriptsize (10)} & 0.04 {\scriptsize (15)} & $-$0.07 {\scriptsize (25)} \\
MUSIQ-AVA~\cite{ke2021musiq} & 0.01 {\scriptsize (20)} & 0.13 {\scriptsize (10)} & 0.29 {\scriptsize (5)} & 0.27 {\scriptsize (0)} \\
MUSIQ-PaQ2PiQ~\cite{ke2021musiq} & 0.14 {\scriptsize (20)} & 0.20 {\scriptsize (20)} & 0.17 {\scriptsize (10)} & $-$0.12 {\scriptsize (20)} \\
MUSIQ-SPAQ~\cite{ke2021musiq} & 0.04 {\scriptsize (10)} & 0.07 {\scriptsize (15)} & 0.03 {\scriptsize (15)} & $-$0.12 {\scriptsize (30)} \\
NIMA~\cite{talebi2018nima} & \cmark\ $-$0.23 {\scriptsize (50)} & $-$0.09 {\scriptsize (30)} & $-$0.06 {\scriptsize (10)} & \cmark\ $-$0.03 {\scriptsize (25)} \\
NIMA-KONIQ~\cite{talebi2018nima} & \cmark\ $-$0.21 {\scriptsize (55)} & $-$0.19 {\scriptsize (45)} & $-$0.22 {\scriptsize (35)} & \cmark\ $-$0.27 {\scriptsize (75)} \\
NIMA-SPAQ~\cite{talebi2018nima} & \cmark\ $-$0.07 {\scriptsize (30)} & 0.00 {\scriptsize (25)} & 0.01 {\scriptsize (0)} & $-$0.13 {\scriptsize (45)} \\
NIMA-VGG16-AVA~\cite{talebi2018nima} & \cmark\ $-$0.13 {\scriptsize (60)} & \cmark\ $-$0.05 {\scriptsize (35)} & \cmark\ $-$0.13 {\scriptsize (40)} & \cmark\ $-$0.09 {\scriptsize (45)} \\
NIQE~\cite{mittal2012making} & $-$0.14 {\scriptsize (0)} & 0.24 {\scriptsize (0)} & 0.15 {\scriptsize (0)} & 0.12 {\scriptsize (0)} \\
NIQE (MATLAB)~\cite{mittal2012making} & $-$0.17 {\scriptsize (0)} & 0.19 {\scriptsize (0)} & 0.23 {\scriptsize (0)} & 0.24 {\scriptsize (0)} \\
NMFC~\cite{chen2023underwater} & $-$0.15 {\scriptsize (15)} & $-$0.10 {\scriptsize (10)} & 0.12 {\scriptsize (25)} & \cmark\ $-$0.06 {\scriptsize (40)} \\
NR-UIQA~\cite{yi2024no} & $-$0.18 {\scriptsize (10)} & $-$0.27 {\scriptsize (35)} & \cmark\ $-$0.22 {\scriptsize (60)} & \cmark\ $-$0.21 {\scriptsize (80)} \\
NRQM~\cite{ma2017learning} & 0.11 {\scriptsize (5)} & 0.24 {\scriptsize (10)} & 0.07 {\scriptsize (15)} & 0.03 {\scriptsize (30)} \\
NUIQ~\cite{jiang2022underwater} & $-$0.26 {\scriptsize (5)} & $-$0.06 {\scriptsize (5)} & 0.03 {\scriptsize (15)} & \cmark\ $-$0.04 {\scriptsize (30)} \\
PaQ-2-PiQ~\cite{ying2020patches} & 0.03 {\scriptsize (25)} & $-$0.02 {\scriptsize (25)} & $-$0.20 {\scriptsize (30)} & \cmark\ $-$0.35 {\scriptsize (60)} \\
PI~\cite{blau20182018} & $-$0.01 {\scriptsize (0)} & 0.27 {\scriptsize (5)} & 0.14 {\scriptsize (20)} & 0.10 {\scriptsize (20)} \\
PIQE~\cite{venkatanath2015blind} & $-$0.59 {\scriptsize (50)} & $-$0.23 {\scriptsize (10)} & \cmark\ $-$0.23 {\scriptsize (75)} & \cmark\ $-$0.22 {\scriptsize (100)} \\
Q-Align~\cite{wu2023q} & \cmark\ $-$0.68 {\scriptsize (70)} & $-$0.51 {\scriptsize (45)} & $-$0.51 {\scriptsize (10)} & \cmark\ $-$0.41 {\scriptsize (35)} \\
QualiCLIP~\cite{agnolucci2024quality} & \cmark\ $-$0.11 {\scriptsize (50)} & \cmark\ $-$0.01 {\scriptsize (30)} & $-$0.12 {\scriptsize (35)} & $-$0.28 {\scriptsize (45)} \\
QualiCLIP+~\cite{agnolucci2024quality} & 0.35 {\scriptsize (0)} & 0.38 {\scriptsize (5)} & 0.28 {\scriptsize (5)} & 0.00 {\scriptsize (15)} \\
QualiCLIP+-CLIVE~\cite{agnolucci2024quality} & 0.15 {\scriptsize (25)} & \cmark\ $-$0.01 {\scriptsize (25)} & \cmark\ $-$0.26 {\scriptsize (65)} & \cmark\ $-$0.36 {\scriptsize (80)} \\
QualiCLIP+-FLIVE~\cite{agnolucci2024quality} & 0.27 {\scriptsize (0)} & 0.24 {\scriptsize (5)} & 0.32 {\scriptsize (5)} & 0.11 {\scriptsize (10)} \\
QualiCLIP+-SPAQ~\cite{agnolucci2024quality} & \cmark\ $-$0.03 {\scriptsize (60)} & \cmark\ $-$0.04 {\scriptsize (40)} & \cmark\ $-$0.17 {\scriptsize (60)} & \cmark\ $-$0.34 {\scriptsize (60)} \\
QUIQ~\cite{karimi2025quiq} & 0.12 {\scriptsize (10)} & 0.06 {\scriptsize (5)} & 0.27 {\scriptsize (5)} & 0.18 {\scriptsize (5)} \\
RALI~\cite{zhao2026reasoning} & \cmark\ $-$0.50 {\scriptsize (95)} & \cmark\ $-$0.36 {\scriptsize (55)} & \cmark\ $-$0.44 {\scriptsize (50)} & \cmark\ $-$0.38 {\scriptsize (75)} \\
TOPIQ-IAA~\cite{chen2024topiq} & \cmark\ $-$0.43 {\scriptsize (70)} & $-$0.20 {\scriptsize (30)} & $-$0.07 {\scriptsize (15)} & $-$0.05 {\scriptsize (25)} \\
TOPIQ-IAA R50~\cite{chen2024topiq} & \cmark\ $-$0.07 {\scriptsize (50)} & 0.03 {\scriptsize (25)} & 0.10 {\scriptsize (15)} & $-$0.02 {\scriptsize (10)} \\
TOPIQ-NR~\cite{chen2024topiq} & 0.43 {\scriptsize (0)} & 0.34 {\scriptsize (5)} & 0.32 {\scriptsize (10)} & 0.13 {\scriptsize (10)} \\
TOPIQ-NR-FLIVE~\cite{chen2024topiq} & \cmark\ $-$0.03 {\scriptsize (20)} & 0.06 {\scriptsize (15)} & 0.14 {\scriptsize (5)} & $-$0.02 {\scriptsize (15)} \\
TOPIQ-NR-SPAQ~\cite{chen2024topiq} & 0.15 {\scriptsize (30)} & 0.09 {\scriptsize (20)} & \cmark\ $-$0.13 {\scriptsize (55)} & \cmark\ $-$0.40 {\scriptsize (75)} \\
TReS~\cite{golestaneh2022no} & 0.49 {\scriptsize (0)} & 0.42 {\scriptsize (10)} & 0.31 {\scriptsize (5)} & 0.13 {\scriptsize (15)} \\
TReS-FLIVE~\cite{golestaneh2022no} & 0.02 {\scriptsize (25)} & \cmark\ $-$0.10 {\scriptsize (40)} & $-$0.08 {\scriptsize (20)} & $-$0.22 {\scriptsize (30)} \\
Twice-Mixing~\cite{fu2022twice} & \cmark\ $-$0.37 {\scriptsize (60)} & $-$0.32 {\scriptsize (30)} & $-$0.27 {\scriptsize (10)} & $-$0.30 {\scriptsize (30)} \\
UCIQE~\cite{yang2015underwater} & 0.24 {\scriptsize (0)} & 0.36 {\scriptsize (0)} & 0.26 {\scriptsize (5)} & 0.25 {\scriptsize (0)} \\
UCIQE (MATLAB code)~\cite{wong2020uciqecode} & 0.57 {\scriptsize (0)} & 0.56 {\scriptsize (0)} & 0.43 {\scriptsize (0)} & 0.29 {\scriptsize (0)} \\
UICM~\cite{panetta2015human} & 0.18 {\scriptsize (0)} & 0.37 {\scriptsize (0)} & 0.32 {\scriptsize (0)} & 0.38 {\scriptsize (0)} \\
UIConM~\cite{panetta2015human} & $-$0.54 {\scriptsize (35)} & $-$0.50 {\scriptsize (35)} & \cmark\ $-$0.30 {\scriptsize (90)} & \cmark\ $-$0.23 {\scriptsize (65)} \\
UIQAVSI~\cite{chen2025underwater} & $-$0.22 {\scriptsize (40)} & $-$0.12 {\scriptsize (30)} & \cmark\ $-$0.02 {\scriptsize (45)} & \cmark\ $-$0.21 {\scriptsize (80)} \\
UIQM~\cite{panetta2015human} & $-$0.11 {\scriptsize (5)} & $-$0.07 {\scriptsize (0)} & \cmark\ $-$0.05 {\scriptsize (55)} & \cmark\ $-$0.06 {\scriptsize (50)} \\
UISM~\cite{panetta2015human} & 0.16 {\scriptsize (0)} & 0.14 {\scriptsize (10)} & 0.08 {\scriptsize (20)} & $-$0.03 {\scriptsize (30)} \\
UNIQUE~\cite{zhang2021uncertainty} & 0.22 {\scriptsize (0)} & 0.01 {\scriptsize (15)} & $-$0.18 {\scriptsize (20)} & $-$0.27 {\scriptsize (25)} \\
URanker~\cite{guo2023underwater} & $-$0.06 {\scriptsize (20)} & $-$0.15 {\scriptsize (25)} & 0.01 {\scriptsize (5)} & $-$0.05 {\scriptsize (30)} \\
VisualQuality-R1~\cite{wu2026visualquality} & \cmark\ $-$0.69 {\scriptsize (45)} & $-$0.56 {\scriptsize (25)} & $-$0.62 {\scriptsize (30)} & $-$0.40 {\scriptsize (15)} \\
WaDIQaM-NR~\cite{bosse2017deep} & 0.57 {\scriptsize (0)} & 0.58 {\scriptsize (0)} & 0.55 {\scriptsize (0)} & 0.46 {\scriptsize (0)} \\
\bottomrule
\end{longtable}

}

\end{document}